\documentclass{article}

\PassOptionsToPackage{sort&compress}{natbib}
\usepackage{natbib}

\usepackage[preprint]{neurips_2022}

\usepackage{hyperref}
\usepackage{xcolor}
\hypersetup{
    colorlinks,
    linkcolor={red!50!black},
    citecolor={blue!50!black},
    urlcolor={blue!80!black}
}

\usepackage[textwidth=2.0cm, textsize=tiny]{todonotes} % for writting

\usepackage[normalem]{ulem} % strikethrough

\usepackage[utf8]{inputenc} % allow utf-8 input
\usepackage[T1]{fontenc}    % use 8-bit T1 fonts
\usepackage{hyperref}       % hyperlinks
\usepackage{url}            % simple URL typesetting
\usepackage{booktabs}       % professional-quality tables
\usepackage{amsfonts}       % blackboard math symbols
\usepackage{nicefrac}       % compact symbols for 1/2, etc.
\usepackage{microtype}      % microtypography
\usepackage{xcolor}         % colors
\usepackage{mathtools}
\usepackage{amsmath}
\usepackage{amsthm}
\usepackage{amssymb}

\usepackage{notation}
\usepackage{multicol}

\title{Fair learning with Wasserstein barycenters for non-decomposable performance measures}

\author{%
Solenne Gaucher$^*$\\
Université Paris-Saclay, CNRS\\
      Laboratoire de mathématiques d’Orsay\\
\And
Nicolas Schreuder$^*$\\
 MaLGa, DIBRIS\\
 Università di Genova
\And
Evgenii Chzhen
   \\
   Université Paris-Saclay, CNRS\\
      Laboratoire de mathématiques d’Orsay\\
}

\begin{document}

\maketitle

\begin{abstract}
This work provides several fundamental characterizations of the optimal classification function under the demographic parity constraint. In the awareness framework, akin to the classical unconstrained classification case, we show that maximizing accuracy under this fairness constraint is equivalent to solving a corresponding regression problem followed by thresholding at level $1/2$. We extend this result to linear-fractional classification measures (e.g., ${\rm F}$-score, AM measure, balanced accuracy, etc.), highlighting the fundamental role played by the regression problem in this framework.
Our results leverage recently developed connection between the demographic parity constraint and the multi-marginal optimal transport formulation. Informally, our result shows that the transition between the unconstrained problems and the fair one is achieved by replacing the conditional expectation of the label by the solution of the fair regression problem.
Finally, leveraging our analysis, we demonstrate an equivalence between the awareness and the unawareness setups in the case of two sensitive groups.
\end{abstract}

\section{Introduction}
%The rapid and wide deployment of decision-making algorithms in the last couple of years 

Our\blfootnote{$^{*}$ Equal contribution} experience of life is increasingly and insidiously being influenced by algorithmic predictions. It is now well accepted that such predictions might replicate or even amplify societal biases and discrimination because of machine learning algorithms’ training process \citep{barocas-hardt-narayanan}.
A key difficulty in overcoming the effect of those biases is the lack of a precise understanding of how statistical algorithms make predictions: these algorithms are often designed to minimize a user-specified data-dependent loss and yield a highly complex prediction rule, leaving practitioners---and theoreticians---unable to understand and explain the issued predictions.
Our goal is to provide a sound and simple mathematical characterization of the prediction process in the presence of fairness constraints.

In this paper we study the demographic parity fairness constraint~\citep{calders2009building,barocas-hardt-narayanan} in the \emph{awareness} framework---allowing the prediction rules to explicitly take the sensitive attribute as an input. Even though this constraint is relatively well understood from algorithmic perspective in both classification~\citep{Agarwal_Beygelzimer_Dubik_Langford_Wallach18,menon2018cost,zeng2022bayes,schreuder2021classification,yang2020fairness,jiang2019wasserstein,silvia2020general,feldman2015certifying,gordaliza2019obtaining} and regression~\citep{chzhen2020fair,chzhen2020fairTV,gouic2020price,jiang2019wasserstein,agarwal2019fair,chiappa2021fairness}, the connection between the two setups remains opaque. The main goal of the current paper is to unveil it.

In contrast, in the traditional unconstrained learning setup, the relation between classification and its regression counterpart is well understood and can be found in all standard books on the subject~\citep[see,  e.g.,][]{hastie2009elements,devroye2013probabilistic,james2013introduction,mohri2018foundations}. For instance, the most standard result illustrating this connection states that if $\eta$ minimizes the squared risk, the classifier $g^*(\cdot) = \ind{\eta(\cdot) \geq 1/2}$ minimizes the misclassification error.
Such results form the first building block of many theoretical and practical studies~\citep[see,  e.g.,][]{Audibert_Tsybakov07,Yang99,massart2006risk,biau2008consistency}. More recently, the connection between regression and classification was pushed even further. For instance, replacing the misclassification error by the ${\rm F}$-score \citep{van1974foundation,Chinchor92}, \citet{Zhao_Edakunni_Pocock_Brown13} showed that the solution of the associated regression problem still plays a crucial role as an ${\rm F}$-score maximizer can be obtained by properly thresholding the solution of the regression problem. Moreover, a recent thread of results establish this fundamental relation for a large variety of performance measures including AM measure, the Jaccard similarity coefficient, and G-mean to name a few~\citep{menon2013statistical, koyejo2014consistent,Koyejo_Natarajan_Ravikumar_Dhillon15,Yan_Koyejo_Zhong_Ravikumar18}. Again, akin to the standard minimization of misclassification error problem, all these developments led to many theoretical and practical advances~\citep[see,  e.g.,][]{jasinska2016extreme,chzhen2020optimal,narasimhan2015optimizing,kotlowski2016surrogate,bascol2019cost, boughorbel2017optimal}.
Interestingly, some works that consider group fairness constraints actually report ${\rm F}$-score as a performance measure in their empirical studies without actually tailoring an algorithm to optimize it directly~\citep[see,  e.g.,][]{BiswasR20, BiswasR21, abs-2207-03277, wang2021analyzing,dablain2022towards,wick2019unlocking}.
A possible cause of this is the absence of characterization of fair (${\rm F}$-score) optimal classifiers in the fairness literature. In this paper we fill this gap for the demographic parity constraint and a large class of performance measures.

Literature that treats group fairness notions is typically distinguished by two features: exact notion of fairness and access to the sensitive attribute at prediction time. While this work considers only demographic parity, we discuss both awareness and unawareness setups---allowing or not the access to the sensitive attribute at prediction time respectively. Unlike the case of awareness, in which a significant understanding has been achieved from theoretical perspective, the case of unawareness remains opaque with contributions mainly focusing on algorithmic constructions~\cite[see  e.g.,][]{Agarwal_Beygelzimer_Dubik_Langford_Wallach18,agarwal2019fair,oneto2019general,Donini17, narasimhan2018learning}. A notable work of~\cite{lipton2018does} puts forward several empirical evidences highlighting critical issues connected of the unawareness framework. Our work makes a step towards a more explicit and transparent description of the optimal classifier under the demographic parity constraint with unawareness by introducing a simple theoretical reduction scheme to the awareness setup for binary protected attribute. Consequently, our results support theoretically the empirical claims made by~\cite{lipton2018does}.

\paragraph{Contributions.} The goal of this work is to establish a link between the regression and classification problems under the demographic parity constraint. 
We make the following contributions to the study of algorithmic fairness:
\begin{enumerate}
\itemsep0em
    \item We show that, under mild assumptions, if $f^*$ minimizes the squared risk under the demographic parity constraint, then $\ind{f^* \geq 1/2}$ minimizes the probability of misclassification under the same constraint. 
    \item We extend the above result to a large family of performance measures introduced in~\cite{Koyejo_Natarajan_Ravikumar_Dhillon15} for unconstrained classification. 
    \item In the case of a binary sensitive attribute, we provide a simple reduction scheme that transforms, in a optimal way, the \emph{unawareness} setup into the \emph{awareness} one.
\end{enumerate}
The first two contributions show the fundamental role played by regression in the context of demographic parity constraint and are built using basic tools from univariate optimal transport theory. As an interesting consequence of our analysis, we show that the notion of strong demographic parity introduced by~\cite{jiang2019wasserstein} is equivalent to the usual demographic parity when a performance measure is minimized. The latter indicates that the post-hoc or the downstream threshold will never harm the demographic parity constraint. The last contribution constitutes a step towards the theoretical treatment of the unawareness setup---a problem that still remains open. Importantly, even though our results are stated in the fair learning setting, they imply new results in the general learning setting. In particular, our results allow to obtain the characterization of the optimal unconstrained classifier for a large class of classification performance measures.

\section{Problem setup}\label{sec:problem_setup}
Consider a triplet $(\bX, S, Y) \in \class{X} \times [K] \times \{0, 1\}$, following some joint distribution $\Prob$, consisting of the nominally non-sensitive and sensitive features, and the label, respectively. Classifiers are functions of the form $g : \class{X} \times [K] \mapsto \{0, 1\}$ and score functions take the form $f : \class{X} \times [K] \mapsto [0, 1]$. The set of all classifiers is denoted by $\class{G}$ and the set of all score functions is denoted by $\class{F}$.
Before proceeding let us introduce additional notation that is related to the unknown distribution $\Prob$. We set $\eta(\bX, S) \triangleq \Exp[Y \mid \bX, S]$ and recall that $\eta$ minimizes the squared risk without any constraint. For each $s \in [K]$, we define $p_s \triangleq \Prob(S = s)$. The central object of this work is the optimal fair score function, defined as:
\begin{highlighted}
\begin{align}
        \label{eq:def_fair_score}
        f^* \in \argmin_{f \in \class{F}} \enscond{\Exp(Y - f(\bX, S))^2}{f(\bX, S) \independent S}\enspace.
\end{align}
\end{highlighted}
An explicit expression for $f^*$ under standard assumptions was derived in~\citep{chzhen2020fair, gouic2020price} using the univariate optimal transport theory and the reduction of the problem in Eq.~\eqref{eq:def_fair_score} to a multi-marginal optimal transport formulation.
In particular, they showed that, under mild assumptions, there is a one-to-one correspondence between the problem in Eq.~\eqref{eq:def_fair_score} and the problem
\begin{align*}
    \min_{\nu \in \class{P}_2(\bbR)} \sum_{s = 1}^K p_s {\sf W}_2^2\parent{\Law(\eta(\bX, S) \mid S = s),\, \nu}\enspace,
\end{align*}
where ${\sf W}_2$ is the Wasserstein-2 distance \citep[Definition 6.1]{villani2009optimal} and $\class{P}_2(\bbR)$ denotes the space of univariate probability measures with finite second moment. Denoting by $\nu^{\star}$ the solution of the above problem, it was shown that
\begin{align*}
    f^*(\bx, s) = T_{\Law(\eta(\bX, S) \mid S = s) \to \nu^{\star}} \, \bigg(\eta(\bx, s)\bigg)\enspace,
\end{align*}
where $T_{\Law(\eta(\bX, S) \mid S = s) \to \nu^{\star}}$ is the optimal transport map from $\Law(\eta(\bX, S) \mid S = s)$ to $\nu^{\star}$. 
Up until now, unlike in the regression setting, it was not clear if a direct link between optimal transport and the fair binary classification problem existed--or even made sense. Our work shows that such a connection exists and that it is fundamental.%/meaningful. Up until now, a clear link between optimal transport and the fair binary classification problem has been missing. The main goal of this work is to provide one.

\paragraph{Notation.} Given a real-valued function $f : \class{X} \times [K] \to \bbR$, we denote by $\mu_s(f)$ the univariate measure defined for all $A \subset \bbR$ as $\mu_s(f)(A) \triangleq \Prob(f(\bX, S) \in A \mid S = s)$. For any univariate measure $\mu$, we denote by $F_{\mu}$ its cumulative distribution, and by $F_{\mu}^{-1}$ its quantile function, given by $F_{\mu}^{-1}(p) \triangleq \min\{x \,:\, \mu((-\infty, x]) \geq p\}$. For any $x \in \bbR$ we set $(x)_+ \triangleq \max\{x,\, 0\}$. For any probability measure $\mu$ on $\class{X}$ and a function $f : \class{X} \to \bbR$, we denote by $f \sharp \mu$, the image measure of $\mu$.

\section{The misclassification risk: a warm-up}
In this section, we begin by tackling the classical minimization of the misclassification risk problem and highlight the main novelties and advances with respect to previous works. To this end, we consider the following optimal (in terms of the misclassification risk) fair classifier
\begin{highlighted}
\begin{align}
    \label{eq:def_fair_classif}
    g^* \in \argmin_{g \in \class{G}} \enscond{\Prob(Y \neq g(\bX, S))}{g(\bX, S) \independent S}\enspace.
\end{align}
\end{highlighted}
We work under the following assumption.
\begin{assumption}
\label{ass:continuity0}
For every $s \in [K]$, assume that $\Law(\eta(\bX, S) \mid S = s)$ is continuous and supported on an interval.
\end{assumption}
A slightly modified version of the above was used in the context of fairness in~\citep{chzhen2020fair,chzhen2020fairTV,gouic2020price,jiang2019wasserstein} and also also in the classical unconstrained classification with generalized performance measures~\citep{Yan_Koyejo_Zhong_Ravikumar18}. In Section~\ref{sec:unified_proof}, we relax the above assumption and provide a proof that unifies the awareness case considered just below with the unawareness case presented in Section~\ref{sec:unwareness}, Theorem~\ref{thm:optimal_DP_unawareness0}.

The first warm-up result is reminiscent of those recently obtained by~\citep{zeng2022bayes,schreuder2021classification}. The proof based on the $\min\max$ duality and is very similar to the classical Neyman-Pearson lemma. While it does not allow to immediately reach our goals, it gives several fundamental insights that were already invoked in previous works on the demographic parity constraint~\citep{lipton2018does,hardt2016equality}.
\begin{theorem}
\label{thm:optimal_DP_0}
Let Assumption~\ref{ass:continuity0} be satisfied. Then, an optimal fair classifier $g^* : \class{X} \times [K] \to \{0, 1\}$ defined in Eq.~\eqref{eq:def_fair_classif} can be expressed for all $(\bx, s) \in \class{X} \times [K]$ as
\begin{align*}
     &g^*(\bx, s) = \ind{2\eta(\bx, s) - 1 \geq \frac{\lambda_s^*}{p_s}}\enspace,
\end{align*}
where $\blambda^* = (\lambda_1^*, \ldots, \lambda_K^*)^\top \in \bbR^K$ is a solution of
\begin{align*}
     \min_{\blambda \in \bbR} \enscond{\Exp\parentsq{\abs{ 2\eta(\bX, S) - 1 - \frac{\lambda_S}{p_S}}}}{\Exp\left[\frac{\lambda_S}{p_S}\right] = 0}\enspace.
\end{align*}
\end{theorem}
The main takeaway message from the above theorem is: under the stated assumption, the optimal fair classifier can be derived as a group-wise thresholding of the regression function $\eta$, with thresholds eventually depending on the sensitive groups. For a similar statement without the continuity assumption, we refer the reader to~\citet{zeng2022bayes} who derived optimal randomized classifiers using the Neyman-Pearson lemma.
Let us now provide a novel characterization of an optimal fair classifier.
\begin{mytheo}{Wasserstein based fair optimal classifier}{equivalence}
    Let Assumption~\ref{ass:continuity0} be satisfied. Then, an optimal fair classifier $g^* : \class{X} \times [K] \to \{0, 1\}$ defined in Eq.~\eqref{eq:def_fair_classif} can be expressed for all $(\bx, s) \in \class{X} \times [K]$ as
\begin{align*}
    g^*(\bx, s) = \ind{f^*(\bx, s) \geq 1/2}\text{ with $f^*$ being defined in~\eqref{eq:def_fair_score}}\enspace,
\end{align*}
\end{mytheo}

% \begin{theorem}[Wasserstein based fair optimal classifier]
% \label{thm:equivalence}
% Let Assumption~\ref{ass:continuity0} be satisfied. Then, an optimal fair classifier $g^* : \class{X} \times [K] \to \{0, 1\}$ defined in Eq.~\eqref{eq:def_fair_classif} can be expressed for all $(\bx, s) \in \class{X} \times [K]$ as
% \begin{align*}
%     g^*(\bx, s) = \ind{f^*(\bx, s) \geq 1/2}\enspace.
% \end{align*}
% \end{theorem}
\paragraph{Discussion.} The above result is instructive on its own---one can solve binary classification under the demographic parity constraint by solving the corresponding regression problem. We recall that~\citep{chzhen2020fair, gouic2020price} built a statistically consistent algorithm for the estimation of the latter. Furthermore, they showed that under the imposed assumptions,
\begin{align*}
    f^*(\bx, s) = \underbrace{\bigg(\sum_{\sigma = 1}^Kp_{\sigma}F^{-1}_{\mu_\sigma(\eta)}\bigg) \circ F_{\mu_s(\eta)}}_{\text{transport to the barycenter}} \circ \, \eta(\bx, s)\enspace.
\end{align*}
%\evg{This is new: below}
\cite{feldman2015certifying} proposed to transport the group-wise distribution of $\eta(\bX, S)$ towards their common barycenter as a disparity removal strategy. Yet, a theoretical justification was missing and this approach remained a heuristic until the work of~\cite{gordaliza2019obtaining} who provided an upper bound on the excess risk in terms of the Wasserstein barycenter objective. 
Later,~\cite{jiang2019wasserstein} relied on the barycenter formulation involving the Earth Mover distance~\citep{rachev1998mass} and showed that a transport-based prediction results in a minimal perturbation post-processing. However, the use of the Earth Mover distance might result in non-uniqueness issues. Our Theorem~\ref{thm:equivalence} gives a complete theoretical justification of the transport based fair classification algorithms. Theorem~\ref{thm:fair_optimal_LF} in Section~\ref{sec:non_decomposable} further extends this connection to non-decomposable measures.

Besides, \citet{jiang2019wasserstein} introduced a notion of strong demographic parity, which amounts to taking classifiers $g : \class{X} \times [K] \to \{0, 1\}$ for which there exists a score function $f: \class{X} \times [K] \to [0, 1]$ such that $f(\bX, S) \independent S$ and $g(\bx, s) = \ind{f(\bx, s) \geq 1/2}$. This notion was later used in~\citep{silvia2020general,chiappa2021fairness}.
Theorem~\ref{thm:equivalence} implies that the optimal classifier under the demographic parity constraint satisfies, \emph{an a priori more restrictive} fairness notion---the strong demographic parity. Indeed, any classifier that satisfies strong demographic parity is demographic parity fair. Hence, we have deduced the equivalence between the two definitions at the optimum.
The notion of strong demographic parity introduced by~\citet{jiang2019wasserstein} can be seen in a downstream or post-hoc settings. That is, the learner first tries to fit a score function and only after a particular threshold is selected in a potentially non-stationary way. Strong demographic parity implies that \emph{any} threshold selection made by the learner will yield a fair classifier. In that sense, our results show that building a score function via an optimal fair regression function is optimal for misclassification risk and, as we see in Section~\ref{sec:non_decomposable}, for many other classification measures. Below we provide a simple proof of Theorem~\ref{thm:equivalence}.

\begin{figure}[t!]
    \centering
    \includegraphics[width=0.325\textwidth]{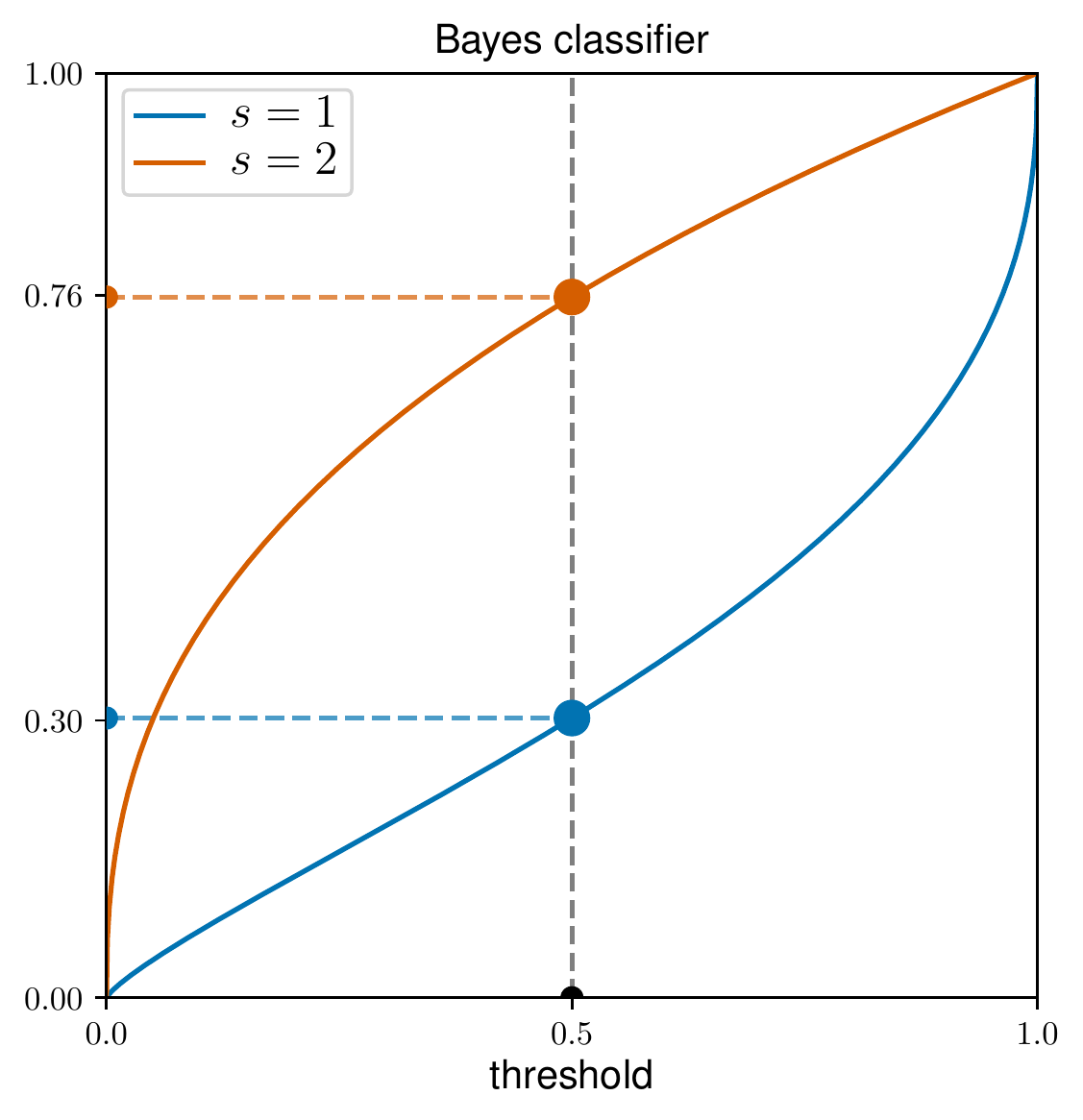}
    \includegraphics[width=0.325\textwidth]{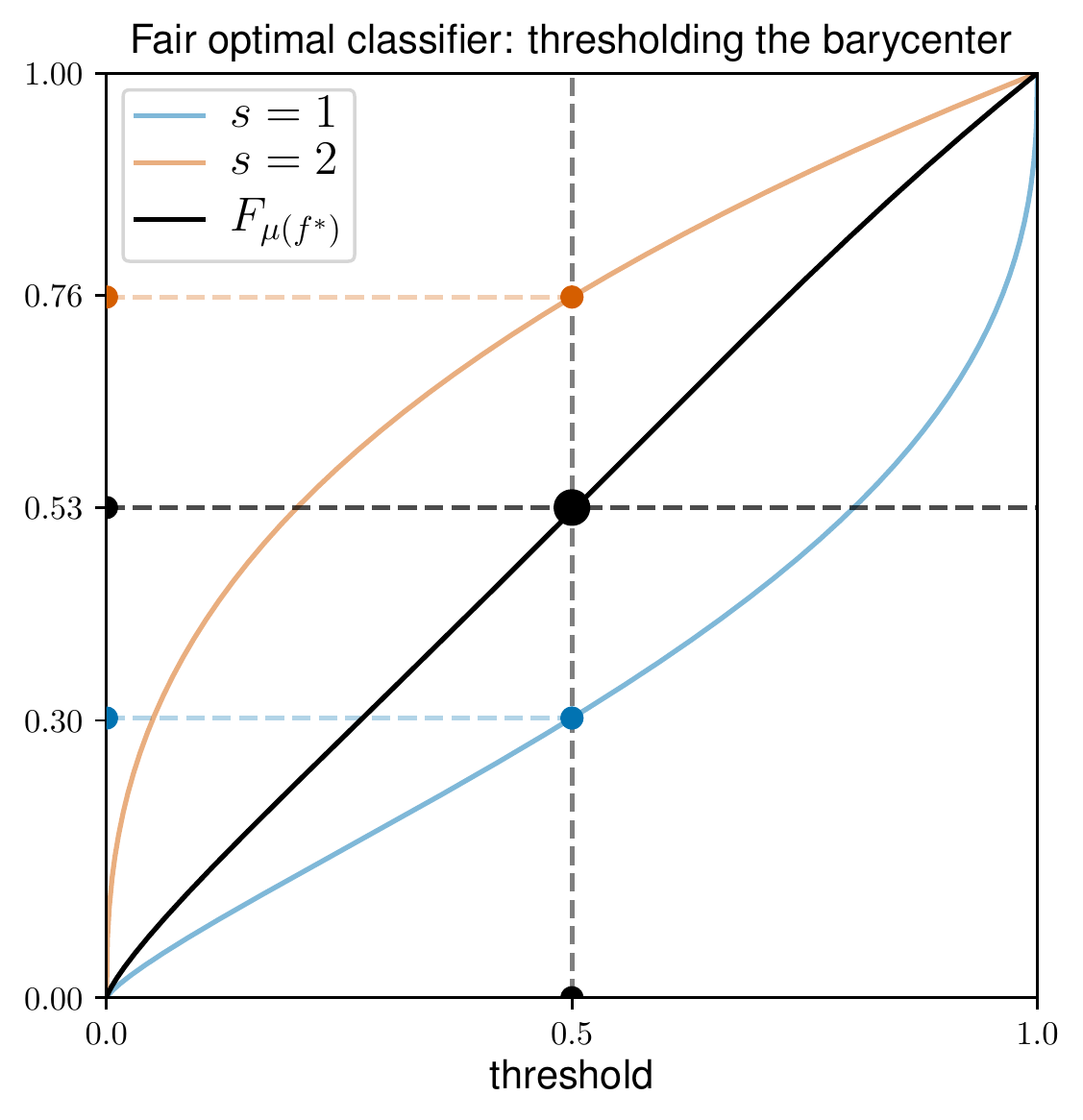}
    \includegraphics[width=0.325\textwidth]{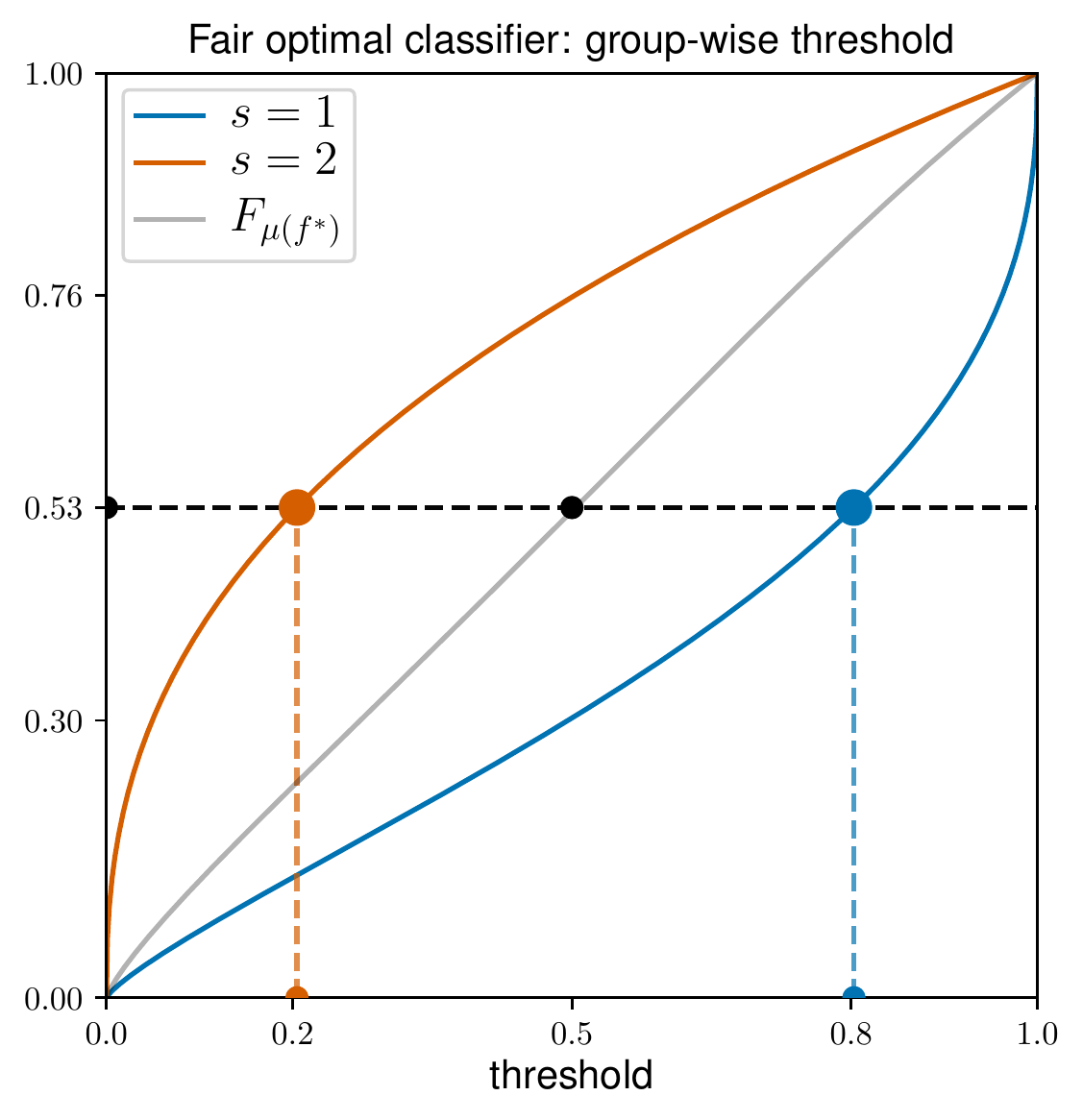}
    \caption{Illustration of Bayes and fair optimal classifiers. \texttt{Left}: group-wise cumulative distributions of $\eta(\bX, S)$---the threshold is $.5$; \texttt{Middle}: Illustration of Theorem~\ref{thm:equivalence}---black solid line corresponds to $F_{\mu(f^*)}$; \texttt{Right}: illustration of group-wise thresholds that correspond to the fair optimal classifier.}
    \label{fig:main}
\end{figure}

\begin{proof}[Proof of Theorem~\ref{thm:equivalence}]

Theorem~\ref{thm:optimal_DP_0} implies that under Assumption~\ref{ass:continuity0} the optimal classifier is of the form $g^*(x,s) = \ind{\eta(\bx, s) \geq \beta_s^*}$ for some $\bbeta^* = (\beta_s^*)_{s \in [K]} \in \bbR^K$. It follows from~\cite[Lemma 21.1(iv)]{van2000asymptotic} and Assumption~\ref{ass:continuity0} that $\eta(\bx,s) = F_{\mu_s(\eta)}^{-1}\circ F_{\mu_s(\eta)}(\eta(\bx, s))$ for almost all $\bx \in \bbR^d$ w.r.t. $\Prob_{\bX \mid S = s}$.
Thus, it is sufficient to look at the classifiers of the form
\[g(\bx, s) = \ind{F_{\mu_s(\eta)}^{-1}\circ F_{\mu_s(\eta)}(\eta(\bx, s)) \geq \beta_s}\enspace,\] 
or, equivalently, at $g(\bx, s) = \ind{F_{\mu_s(\eta)}(\eta(\bx, s)) \geq F_{\mu_s(\eta)}(\beta_s)}$~\cite[Lemma 21.1(i)]{van2000asymptotic}. Now, the inverse transform theorem states that under Assumption \ref{ass:continuity0}, $F_{\mu_s(\eta)}^{-1}(U)$ has the same distribution as $\eta(\bX, S)$ conditionally on $S = s$, for $U$ uniformly distributed on $(0,1)$. Then,
\[
\mathbb{P}\left(g(\bX, S) = 1 \mid S = s\right) =  \mathbf{P}\left(F_{\mu_s(\eta)}\circ F_{\mu_s(\eta)}^{-1}(U) \geq F_{\mu_s(\eta)}(\beta_s) \right) =  1 - F_{\mu_s(\eta)}(\beta_s)\enspace,
\]
where we have used that $F_{\mu_s(\eta)}\circ F_{\mu_s(\eta)}^{-1}(u) = u$ for all $u \in (0,1)$ ~\cite[Lemma 21.1(ii)]{van2000asymptotic}. Thus, $g$ verifies the DP constraint if and only if $F_{\mu_s(\eta)}(\beta_s)$ does not depend on $s$. Denoting by $\gamma$ this constant, we find that the optimal fair classifier must be of the form $g(\bx, s) = \ind{F_{\mu_s(\eta)}(\eta(\bx, s))\geq \gamma}$.
% \[
% g(\bx, s) = \ind{F_{\mu_s(\eta)}(\eta(\bx, s)) \geq \gamma}\enspace.\]
The risk of any such classifier is given by
\begin{align}
\label{eq:solenne_0}
\risk(g) = \mathbb{E}[Y] + \sum_{s \in [K]}p_s \Exp[\ind{F_{\mu_s(\eta)}(\eta(\bx, s)) \geq \gamma}(1 -2\eta(\bX,s)) \mid S = s]\enspace.
\end{align}
Using again inverse transform theorem, Eq.~\eqref{eq:solenne_0} can be further simplified to the following expression:
\begin{align}
\label{eq:solenne_1}
\risk(g) = \mathbb{E}[Y] + \sum_{s \in [K]}p_s \underset{0}{\overset{1}{\int}}\ind{F_{\mu_s(\eta)}\circ F_{\mu_s(\eta)}^{-1}(u) \geq \gamma}(1 - 2F_{\mu_s(\eta)}^{-1}(u)) \d u\enspace.
\end{align}
Under Assumption~\ref{ass:continuity0}, $F_{\mu_s(\eta)}\circ F_{\mu_s(\eta)}^{-1}(u) = u$ for all $u \in (0,1)$. Thus, Eq.~\eqref{eq:solenne_1} reduces to
\begin{align*}
\risk(g) = \mathbb{E}[Y] + \underset{\gamma}{\overset{1}{\int}}\sum_{s \in [K]}p_s  (1 - 2F_{\mu_s(\eta)}^{-1}(u))\d u \enspace.
\end{align*}
This function is minimized at $\gamma^*$ which satisfies
\begin{align}\label{eq:def_gamma}
    \bigg(\sum_{s \in [K]}p_sF_{\mu_s(\eta)}^{-1}\bigg)(\gamma^*) = 1/2\enspace,
\end{align}
and the optimal classifier under the demographic parity constraints is given by $g^*(\bx,s) = \ind{F_{\mu_s(\eta)}(\eta(\bx, s)) \geq \gamma^*}$. Taking into account the condition satisfied by $\gamma^*$, we conclude.
\end{proof}
The proof itself is rather instructive and gives rise to the following interpretation.
\begin{myremark}{Ranking interpretation of the fair optimal classification strategy}{Solennes_final}
The proof of Theorem~\ref{thm:equivalence} reveals that the optimal fair classifier can be written as $g^*(\bx, s) = \ind{F_{\mu_s(\eta)}\big(\eta(\bx, s)\big)\geq \gamma^*}$, where $\gamma^*$ is given by \eqref{eq:def_gamma}.
% \textcolor{gray}{Assume that each $F_{\mu_s(\eta)}$ is continuous and strictly increasing, then $\gamma^*$ introduced in the end of the proof of Theorem~\ref{thm:equivalence} equals to
% \begin{align*}
%     \parent{\sum_{s = 1}^K p_s F_{\mu_s(\eta)}^{-1}}^{-1}(1/2)\enspace.
% \end{align*}}
Recall that $q \mapsto \sum_{s = 1}^K p_s F_{\mu_s(\eta)}^{-1}(q)$ is the quantile function of the Wasserstein-2 barycenter of measures $(\mu_s(\eta))_{s \in [K]}$, weighted by $(p_s)_{s \in [K]}$~\cite[see, e.g.,][Section 6.1]{agueh2011barycenters}. 
Thus, denoting this barycenter by $\bar{\mu}(\eta)$, $g^*$ can be alternatively expressed as
\begin{align*}
    g^*(\bx, s) = \ind{F_{\mu_s(\eta)}\big(\eta(\bx, s)\big)\geq F_{\bar{\mu}(\eta)}\big(1/2\big)}\enspace.
\end{align*}
The last display shows that while the thresholds of $\eta$ differ across groups (as per Theorem~\ref{thm:optimal_DP_0}), this threshold sensitive-group independent if viewed from the perspective of group-wise ranking. Putting it simply, if $F_{\bar{\mu}(\eta)}(1/2) = p \in (0, 1)$, then the $(1{-}p) \times 100 \%$ \emph{best} individuals from each group get classified positively. This property reflects the notion of rational ordering~\citep{lipton2018does} that follows from order preservation property of $f^*$~\citep[see][Section 4]{chzhen2020minimax}.
Figure~\ref{fig:main} provides a graphical illustration of the above observations.
\end{myremark}
We note that as in other works explaining a given fairness constraint, we do not argue for or against the policy itself.

% \usepackage{booktabs}
\begin{table}[t!]
\centering
\resizebox{\textwidth}{!}{%
\begin{tabular}{@{}llll@{}}
\toprule
                    & Expression                                                                      & $(\sfn_0, \sfn_1, \sfn_2)$  & $(\sfd_0, \sfd_1, \sfd_2)$    \\ \midrule
Accuracy            & {\small$\Prob(Y = g(\bX, S))$}                                                                & $(1-p^{y=1},\, 2,\, -1)$ & $(1,\, 0,\, 0)$               \\[1.5ex]
${\rm F}_{b}$-score & $\frac{(1+b^2)\Prob(Y = 1,\, g(\bX, S) = 1)}{b^2\Prob(Y = 1) + \Prob(g(\bX, S) = 1)}$ & $(0,\, 1+b^2,\, 0)$         & $(b^2p^{y=1},\, 0,\, 1)$
% AM-measure          & $\frac{p^{y=0}\Prob(Y = 1,\, g(\bX, S)= 1) + p^{y=1}\Prob(Y = 0,\, g(\bX, S)=0)}{2p^{y=1}p^{y=0}}$                                                                                  & $??$                        & $??$                      
\\ [1.5ex]
Jaccard             & $\frac{\Prob(Y = 1,\, g(\bX, S) = 1)}{\Prob(Y = 1,\, g(\bX, S) = 0) + \Prob(g(\bX, S) = 1)}$                                                                                  & $(0,\, 1,\, 0)$                        & $(p^{y=1},\, -1, 1)$                          \\ [1.5ex]
AM-measure                & {\small$\frac{1}{2}\big\{{\small \Prob(g(\bX, S) {=} 0 \mid Y {=} 0) {+} \Prob(g(\bX, S) {=} 1 \mid Y {=} 1)}\big\}$ }                                                                                 & $(\tfrac{1}{2},\, \tfrac{1}{2p^{y=1}} {\small+} \tfrac{1}{2p^{y=0}},\, - \tfrac{1}{2p^{y=0}})$                        & $(1,\, 0,\, 0)$\\ [1.5ex]

Recall                & {\small$\Prob(g(\bX, S) = 1 \mid Y = 1 )$}                                                                                 & $(0,\, 1,\, 0)$                        & $(p^{y=1},\, 0,\, 0)$\\ [1.5ex]
\end{tabular}
}

\caption{Some examples of measure that can be represented by Eq.~\eqref{eq:lin_frac_general_expression}. For more example see~\cite{choi2010survey}. We set for this table $p^{y=1} \triangleq \Prob(Y = 1)$ and $p^{y=0} \triangleq \Prob(Y = 0)$.}
\label{tab:examples}
\end{table}

\section{Non-decomposable performance measures}
\label{sec:non_decomposable}
In this part we extend the analysis of the previous section to a broader class of performance measures, which includes the ${\rm F}$-score, the AM-mean, and the misclassification risk among others. We follow the framework put forward by~\cite{koyejo2014consistent}, who introduced the so-called \emph{linear fractional performance measures}. Formally, given coefficients $(\sfn_0, \sfn_1, \sfn_2) \in \bbR^3$ and $(\sfd_0, \sfd_1, \sfd_2) \in \bbR^3$, the performance of a classifier $g : \class{X} \times [K] \to \{0, 1\}$ is measured by its utility
\begin{align}
    \label{eq:lin_frac_general_expression}
    {\rm U}_{(\sfn, \sfd)}(g) \coloneqq \frac{\sfn_0 + \sfn_1 \Prob(g(\bX, S) = 1, Y = 1) + \sfn_2 \Prob(g(\bX, S) = 1)}{\sfd_0 + \sfd_1 \Prob(g(\bX, S) = 1, Y = 1) + \sfd_2 \Prob(g(\bX, S) = 1)}\enspace.
\end{align}
We denote by $\dom({\rm U}_{(\sfn, \sfd)}) \subset \class{G}$ the set of all classifiers $g : \class{X} \times [K] \to \{0, 1\}$ for which the denominator of ${\rm U}_{(\sfn, \sfd)}$ is non-zero.
It is important to emphasize that both $\sfn$ and $\sfd$ are \emph{allowed} to depend on the unknown distribution of the data $\Prob$ but \emph{not} on the classifier $g$. For instance, the ${\rm F}_1$-score~\citep{van1974foundation} corresponds to the choice $(\sfn_0, \sfn_1, \sfn_2) = (0, 2, 0)$ and $(\sfd_0, \sfd_1, \sfd_2) = (\Prob(Y {=} 1), 0, 1)$. 
We refer to~\citep{choi2010survey} for additional examples of different choices of $(\sfn, \sfd)$ corresponding to different classification performance measures. Recently,~\cite{yang2020fairness} studied linear performance measures in the context of fairness, which essentially corresponds to the special case of the above linear fractional formulation with $(\sfd_0, \sfd_1, \sfd_2)  = (1, 0, 0)$---which, for instance, does not encompass the ${\rm F}_1$-score. In another direction, \cite{celis2019classification} considered linear fractional formulation of \emph{fairness} constraints while optimizing the misclassification risk. However, given the structure of the constraints, this problem can essentially be re-formulated as misclassification risk minimization under linear fairness constraints.\\
As it is common in the literature on generalized performance measures, we view ${\rm U}_{(\sfn, \sfd)}$ as a utility to be \emph{maximized}, contrary to the minimization of the risk viewpoint from the previous section. Thus, our goal is to study%
\begin{highlighted}%
\begin{align}
    \label{eq:def_DP_optimal_LF}
    g^*_{(\sfn, \sfd)} \in \argmax_{g \in \dom({\rm U}_{(\sfn, \sfd)})} \enscond{{\rm U}_{(\sfn, \sfd)}(g)}{g(\bX, S) \independent S}\enspace.
\end{align}
\end{highlighted}%
A remarkable property of linear fractional measures is that the unconstrained maximizer can still be obtained by thresholding the regression function $\eta$. Yet, the threshold in this case might depend on the unknown distribution $\Prob$ and ought to be estimated. Let us provide couple of standard examples.
% \begin{myexample}{Accuracy}{accuracy}
% Consider the problem of maximizing the accuracy:
% \begin{align*}
%     \max_{g \in \class{G}} \Prob(Y = g(\bX, S))\enspace.
% \end{align*}
% Setting $(\sfn_0, \sfn_1, \sfn_2) = (1-\Prob(Y=1), 2, -1)$ and $(\sfd_0, \sfd_1, \sfd_2) = (1, 0, 0)$, we see that the above formulation falls within the considered framework.
% \end{myexample}
\begin{example}
\label{ex:accuracy}
Consider the problem of maximizing the accuracy:
\begin{align*}
    \max_{g \in \class{G}} \Prob(Y = g(\bX, S))\enspace.
\end{align*}
Setting $(\sfn_0, \sfn_1, \sfn_2) = (1-\Prob(Y=1), 2, -1)$ and $(\sfd_0, \sfd_1, \sfd_2) = (1, 0, 0)$, we see that the above formulation falls within the considered framework.
\end{example}
% \begin{myexample}{${\rm F}_1$-score}{F_score}
% Consider the problem of maximizing the ${\rm F}_1$-score:
% \begin{align*}
%     \max_{g \in \class{G}} \frac{2 \Prob(g(\bX, S) = 1, Y = 1)}{\Prob(Y = 1) + \Prob(g(\bX, S) = 1)}\enspace.
% \end{align*}
% \citet{Zhao_Edakunni_Pocock_Brown13} showed that the solution $g^*$ of the above optimization problem can be written as
% \begin{align*}
%     g(\bx, s) = \ind{\eta(\bx, s) \geq \theta^*}\enspace,
% \end{align*}
% where $\theta^* \in [0, 1]$ is the unique solution of $\theta\Prob(Y = 1) = \Exp(\eta(\bX, S) - \theta)_+$.
% \end{myexample}
\begin{example}
\label{ex:F_score}
Consider the problem of maximizing the ${\rm F}_1$-score:
\begin{align*}
    \max_{g \in \class{G}} \frac{2 \Prob(g(\bX, S) = 1, Y = 1)}{\Prob(Y = 1) + \Prob(g(\bX, S) = 1)}\enspace.
\end{align*}
\citet{Zhao_Edakunni_Pocock_Brown13} showed that the solution $g^*$ of the above optimization problem can be written as
\begin{align*}
    g^*(\bx, s) = \ind{\eta(\bx, s) \geq \theta^*}\enspace,
\end{align*}
where $\theta^* \in [0, 1]$ is the unique solution of $\theta\Prob(Y = 1) = \Exp(\eta(\bX, S) - \theta)_+$.
\end{example}
\citet{koyejo2014consistent} pushed further these results demonstrating that the ``thresholding principle'' remains true for the whole family of linear fractional measures. In what follows, we will show that their result is still valid if one replaces $\eta$ by $f^*$---the solution of the fair regression problem. This validity is established in a strong sense, meaning that even the equation (as in Example~\ref{ex:F_score}) determining the threshold is preserved.

\begin{mytheo}{Wasserstein based fair optimal classifier for non-decomposable measures}{fair_optimal_LF}
Let Assumption~\ref{ass:continuity0} be satisfied. Assume that $\sfd \in \bbR^3$ is such that \[\sfd_0 + \min \big\{\min\{\sfd_1,\, 0\} + \sfd_2,\, 0\big\} \geq 0\enspace.\]
Assume that the coefficients $(\sfn, \sfd) \in \bbR^3 \times \bbR^3$ satisfy one of the following mutually exclusive conditions:\\
\noindent\begin{minipage}{.5\linewidth}
\begin{align}
\tag{$\mathtt{C1}$}
 \label{eq:condition1}
    \begin{dcases}
    \sfd_2\sfn_1 > \sfn_2\sfd_1\\
    \frac{\sfn_0\sfd_2 - \sfd_0\sfn_2}{\sfn_2\sfd_1 - \sfd_2\sfn_1} \leq \Prob(Y = 1)\\
    \sfd_0\sfn_1 - \sfn_0\sfd_1 \geq \left(\sfn_0\sfd_2 - \sfd_0\sfn_2\right)_+\\
    \end{dcases}\enspace,
\end{align}
\end{minipage}%
\begin{minipage}{.5\linewidth}
\begin{align}
\tag{$\mathtt{C2}$}
    \label{eq:condition2}
    \begin{dcases}
    \sfd_2\sfn_1 = \sfn_2\sfd_1\\
    \sfn_1\sfd_0 > \sfd_1\sfn_0\\
    \frac{\sfd_0\sfn_2-\sfn_0\sfd_2}{\sfn_0\sfd_1 - \sfd_0\sfn_1} \in [0, 1]
    \end{dcases}\enspace.
\end{align}
\end{minipage}%

Then, $g^*_{(\sfn, \sfd)}$ defined in Eq.~\eqref{eq:def_DP_optimal_LF} can be expressed for all $(\bx, s) \in \class{X} \times [K]$ as
\begin{align}
    \label{eq:fair_optimal_LF}
    g^*_{(\sfn, \sfd)}(\bx, s) = \ind{f^*(\bx, s) \geq \theta^*_{(\sfn, \sfd)}}\enspace,
\end{align}
where $\theta^*_{(\sfn, \sfd)}$ is either the unique solution of
\begin{align}
    \label{eq:fixed_point_LF}
    \Exp\left[(f^*(\bX, S) - \theta)_+\right] = \theta \cdot \left\{\frac{\sfn_0\sfd_1 - \sfd_0\sfn_1}{\sfn_2\sfd_1 - \sfd_2\sfn_1}\right\} + \left\{\frac{\sfn_0\sfd_2 - \sfd_0\sfn_2}{\sfn_2\sfd_1 - \sfd_2\sfn_1} \right\}\enspace,
\end{align}

if $\sfn_2\sfd_1 \neq \sfd_2\sfn_1$ or $\theta^*_{(\sfn, \sfd)} = \tfrac{\sfd_0\sfn_2-\sfn_0\sfd_2}{\sfn_0\sfd_1 - \sfd_0\sfn_1}$ otherwise.
\end{mytheo}

A few comments are in order.
First of all, Theorem~\ref{thm:fair_optimal_LF} states that the pre-cited ‘‘thresholding principle'' still holds for optimizing linear-fractional performance measures under the demographic parity constraint: optimal fair classifiers can be obtained by thresholding the optimal fair regression function $f^*$ at the right threshold level $\theta^*_{(\sfn, \sfd)}$. Moreover, in the case $\sfn_2\sfd_1 = \sfd_2\sfn_1$ an explicit expression is provided, while if $\sfn_2\sfd_1 \neq \sfd_2\sfn_1$ one needs to solve a fixed-point equation to find the optimal threshold. Given that the function defining the fixed-point equation is univariate, monotone and continuous, the bisection method (or any other univariate root-finding method) can be used to obtain an approximation of the optimal threshold up to arbitrary precision.
Finally, since the conditions on the coefficients might seem opaque at first sight, let us argue why they are harmless and meaningful. Intuitively, these conditions specify only two requirements:
\begin{itemize}
    \item The maximization of ${\rm U}_{(\sfn, \sfd)}(g)$ makes sense---the more the classifier align with $Y$ the better. In particular, these conditions exclude $\Prob(Y \neq g(\bX, S))$, whose maximization does not make sense.
    \item The denominator of ${\rm U}_{(\sfn, \sfd)}$ is non-negative.
\end{itemize}
One can verify that all the measures presented in Table~\ref{tab:examples} do indeed satisfy these conditions as well as many other linear fractional performance measures from~\cite{choi2010survey}. We would also like to point out that while the conditions of Theorem~\ref{thm:fair_optimal_LF} are cumbersome, they are easy to check in practice, unlike those given in~\citep{koyejo2014consistent}, who relied on $\sign(\sfn_1 - {\rm U}_{(\sfn, \sfd)}(g^*_{(\sfn, \sfd)})\sfd_1)$. Indeed, to check the latter, one needs to know or estimate the optimal value of ${\rm U}_{(\sfn, \sfd)}$ beforehand, which is not always feasible in practice. In contrast, conditions~\eqref{eq:condition1} and~\eqref{eq:condition2} only involve the known coefficients $(\sfn, \sfd)$. Finally, let us remark that ${\rm U}_{(\sfn, \sfd)}$ = ${\rm U}_{(-\sfn, -\sfd)}$ and both conditions~\eqref{eq:condition1} and~\eqref{eq:condition2} are invariant under the $(\sfn, \sfd) \mapsto (-\sfn, -\sfd)$ transformation. Yet, to fix only one of them, we additionally require $\sfd_0 + \min \big\{\min\{\sfd_1,\, 0\} + \sfd_2,\, 0\big\} \geq 0$, which forces the user to fix the signs of $\sfd$ properly. Let us emphasize that, if $\sfd_0 + \min \big\{\min\{\sfd_1,\, 0\} + \sfd_2,\, 0\big\} > 0$, then $\dom({\sf U}_{(\sfn, \sfd)}) = \class{G}$---the denominator does not zero-out---which is a consequence of Lemma~\ref{lem:denominator_positive}.

\begin{proof}
Let us first show that $\theta^*_{(\sfn, \sfd)}$ exists and unique.
Indeed, the mapping \[\theta \mapsto  \theta \cdot \left\{\frac{\sfn_0\sfd_1 - \sfd_0\sfn_1}{\sfn_2\sfd_1 - \sfd_2\sfn_1}\right\} + \left\{\frac{\sfn_0\sfd_2 - \sfd_0\sfn_2}{\sfn_2\sfd_1 - \sfd_2\sfn_1} \right\} - \Exp\left[(f^*(\bX, S) - \theta)_+\right]\enspace,\] is continuous and monotone increasing on $[0, 1]$ under the specified conditions. On the one hand, for $\theta = 0$ we have $\Exp[f^*(\bX, S)] = \Prob(Y = 1)$~\cite[see][Section 4, item 4 on average stability]{chzhen2020minimax} the above mapping evaluates to $\left\{\tfrac{\sfn_0\sfd_2 - \sfd_0\sfn_2}{\sfn_2\sfd_1 - \sfd_2\sfn_1} \right\} - \Prob(Y = 1) \leq 0$. On the other hand, for $\theta = 1$, it evaluates to  $\left\{\tfrac{\sfn_0\sfd_1 - \sfd_0\sfn_1}{\sfn_2\sfd_1 - \sfd_2\sfn_1}\right\} + \left\{\tfrac{\sfn_0\sfd_2 - \sfd_0\sfn_2}{\sfn_2\sfd_1 - \sfd_2\sfn_1} \right\} \geq 0$. The existence follows from the intermediate value theorem and the uniqueness from monotonicity. The rest of the proof follows from the two lemmas presented below.
\end{proof}

The first lemma is similar to the main result of~\citep{koyejo2014consistent}, while the second one gives an explicit expression for the excess-score of \emph{any} fair classifier. The actual proof technique shares some similarities with the analysis of ${\rm F}_1$-score in~\citep{chzhen2020optimal} who provided an alternative proof to the result of~\cite{Zhao_Edakunni_Pocock_Brown13} recalled in Example~\ref{ex:F_score}.
\begin{mylemma}{Fixed point property}{fixed_at_optim_LF}
Let Assumption~\ref{ass:continuity0} be satisfied.
Let $g^*_{(\sfn, \sfd)}$ be defined in Theorem~\ref{thm:fair_optimal_LF} and assume that $\theta^*_{(\sfn, \sfd)}$ defined in Eq.~\eqref{eq:fixed_point_LF} exists. Then,
\begin{align*}
\begin{dcases}
    {\rm U}_{(\sfn, \sfd)}\left(g^*_{(\sfn, \sfd)}\right) = \frac{\sfn_2 + \theta^*_{(\sfn, \sfd)}\sfn_1}{\sfd_2 + \theta^*_{(\sfn, \sfd)}\sfd_1} &\text{if } \sfn_2\sfd_1 \neq \sfd_2\sfn_1\\
    {\rm U}_{(\sfn, \sfd)}\left(g^*_{(\sfn, \sfd)}\right) = \frac{\sfn_0 + \sfn_1\mathbb{E}{\big( f^*(\bX, S) -\theta^*_{(\sfn, \sfd)}\big)_+}}{\sfd_0 + \sfd_1\mathbb{E}\big( f^*(\bX, S) - \theta^*_{(\sfn, \sfd)}\big)_+} &\text{if } \sfn_2\sfd_1 = \sfd_2\sfn_1
\end{dcases}\enspace.
\end{align*}
\end{mylemma}
% \begin{lemma}
% \label{lem:fixed_at_optim_LF}
% Let Assumption~\ref{ass:continuity0} be satisfied.
% Let $g^*_{(\sfn, \sfd)}$ be defined in Theorem~\ref{thm:fair_optimal_LF} and assume that $\theta^*_{(\sfn, \sfd)}$ defined in Eq.~\eqref{eq:fixed_point_LF} exists. Then,
% \begin{align*}
% \begin{dcases}
%     {\rm U}_{(\sfn, \sfd)}\left(g^*_{(\sfn, \sfd)}\right) = \frac{\sfn_2 + \theta^*_{(\sfn, \sfd)}\sfn_1}{\sfd_2 + \theta^*_{(\sfn, \sfd)}\sfd_1} &\text{if } \sfn_2\sfd_1 \neq \sfd_2\sfn_1\\
%     {\rm U}_{(\sfn, \sfd)}\left(g^*_{(\sfn, \sfd)}\right) = \frac{\sfn_0 + \sfn_1\mathbb{E}{\big( f^*(\bX, S) -\theta^*_{(\sfn, \sfd)}\big)_+}}{\sfd_0 + \sfd_1\mathbb{E}\big( f^*(\bX, S) - \theta^*_{(\sfn, \sfd)}\big)_+} &\text{if } \sfn_2\sfd_1 = \sfd_2\sfn_1
% \end{dcases}\enspace.
% \end{align*}
% \end{lemma}
\begin{proof}
For compactness we drop the subscripts $(\sfn, \sfd)$ in this proof.
Using Lemma ~\ref{lem:f_star_eta_replacing}, we find that
\begin{align*}\mathbb{P}\left(g^*(\bX, S) = 1, Y = 1\right) &= \mathbb{E}\left[f^*(\bX,S)g^*(\bX,S)\right] \\
&= \mathbb{E}\left[\left(f^*(\bX,S)-\theta^*\right)_+\right] + \theta^*\mathbb{E}\left[g^*(\bX,S)\right]\enspace.
\end{align*}
{\textbf Case 1: $\sfn_2\sfd_1 \neq \sfd_2\sfn_1$}.
Combining this result with~\eqref{eq:fixed_point_LF}, we obtain the following expression for ${\rm U}(g^*)$:
\begin{align*}
    \frac{\sfn_0(\sfn_2\sfd_1 - \cancel{\sfd_2\sfn_1}) + \sfn_1\parent{\theta^*(\sfn_0\sfd_1 - \sfd_0\sfn_1) + (\cancel{\sfn_0\sfd_2} - \sfd_0\sfn_2)} + (\sfn_2 + \theta^*\sfn_1)(\sfn_2\sfd_1 - \sfd_2\sfn_1) \Exp[g^*(\bX, S)]}{\sfd_0(\cancel{\sfn_2\sfd_1} - \sfd_2\sfn_1) + \sfd_1\parent{\theta^*(\sfn_0\sfd_1 - \sfd_0\sfn_1) + (\sfn_0\sfd_2 - \cancel{\sfd_0\sfn_2})} + (\sfd_2 + \theta^*\sfd_1)(\sfn_2\sfd_1 - \sfd_2\sfn_1) \Exp[g^*(\bX, S)]}\enspace.
\end{align*}
Factorizing the numerator and denominator by $(\sfn_2 + \theta^*\sfn_1)$ and $(\sfd_2 + \theta^*\sfd_1)$ respectively, the above can be written as
\begin{align*}
    {\rm U}(g^*) = \frac{\sfn_2 + \theta^*\sfn_1}{\sfd_2 + \theta^*\sfd_1} \cdot \frac{(\sfn_0\sfd_1 - \sfd_0\sfn_1) + (\sfn_2\sfd_1 - \sfd_2\sfn_1) \Exp[g^*(\bX, S)]}{(\sfn_0\sfd_1 - \sfd_0\sfn_1) + (\sfn_2\sfd_1 - \sfd_2\sfn_1) \Exp[g^*(\bX, S)]} = \frac{\sfn_2 + \theta^*\sfn_1}{\sfd_2 + \theta^*\sfd_1}\enspace,
\end{align*}
concluding the proof for the first case.\\
{\textbf Case 2: $\sfn_2\sfd_1 = \sfd_2\sfn_1$}. 
In this case, notice that we have
\begin{align*}
    \sfn_1 \theta^* = \frac{\sfn_1 \sfn_2 \sfd_0 - \sfn_0 \sfn_1\sfd_2}{\sfn_0\sfd_1 - \sfn_1 \sfd_0} = \sfn_2 \frac{\sfn_1 \sfd_0 - \sfn_0 \sfd_2}{\sfn_0\sfd_1 - \sfn_1 \sfd_0} = - \sfn_2\enspace,
\end{align*}
and, following the same computations, $\sfd_1 \theta^* = - \sfd_2$.
Plugging the above equalities in the definition of ${\rm U}(g^*)$ yields
\begin{align*}
    {\rm U}(g^*) = \frac{\sfn_0 + \sfn_1 \mathbb{E}{\left(f^*(\bX, S) - \theta^*\right)_+}}{\sfd_0 + \sfd_1\mathbb{E}\left( f^*(\bX, S) - \theta^*\right)_+}\enspace.
\end{align*}
The proof is concluded.
\end{proof}
The next result provides an explicit expression for the excess score of any fair classifier $g$.

\begin{mylemma}{Excess score for fair non-decomposable measures}{excess_score}
Let Assumption~\ref{ass:continuity0} be satisfied.
Let $g^*_{(\sfn, \sfd)}$ be defined as in Theorem~\ref{thm:fair_optimal_LF} and assume that $\theta^* \triangleq \theta^*_{(\sfn, \sfd)}$ defined in Eq.~\eqref{eq:fixed_point_LF} exists.
Let $\bar{\mu}(\eta)$ be the Wasserstein barycenter of measures $\mu_1(\eta), \ldots, \mu_K(\eta)$ weighted by $p_1, \ldots, p_K$, respectively. Define $\beta^*$ as $\beta^* = F_{\bar{\mu}(\eta)}(\theta^*)$.
Let $\class{E}_{(\sfn, \sfd)}(g) \triangleq {\rm U}_{(\sfn, \sfd)}\big(g^*_{(\sfn, \sfd)}\big) - {\rm U}_{(\sfn, \sfd)}\big(g\big)$.
Then, for any classifier $g \in \dom({\rm U}_{(\sfn, \sfd)})$ such that $g(\bX, S) \independent S$, excess score $\class{E}_{(\sfn, \sfd)}(g)$ equals to
\begin{equation*}
     \begin{small}
     \frac{\Exp|{\eta(\bX, S) {-} F_{\mu_S(\eta)}^{-1}(\beta^*)}|\ind{g^*(\bX, S) {\neq} g(\bX, S)}}{{\sfd_0 {+} \sfd_1 \Prob(Y {=} 1,\, g(\bX, S) {=} 1) {+} \sfd_2\Prob(g(\bX, S) {=} 1)}}\cdot
     \begin{dcases}
     \frac{\sfd_2\sfn_1 {-} \sfn_2\sfd_1}{\sfd_2 + \theta^*\sfd_1} & \sfn_2\sfd_1 {\neq} \sfd_2\sfn_1\\
     \frac{\sfn_1\sfd_0 - \sfd_1\sfn_0}{\sfd_0 {+} \sfd_1\Exp(f^*(\bX, S) {-} \theta^*)_+} &\sfn_2\sfd_1 {=} \sfd_2\sfn_1
     \end{dcases}\enspace.
     \end{small}
\end{equation*}
Furthermore, under the conditions on $(\sfn, \sfd)$ specified in Theorem~\ref{thm:fair_optimal_LF}; we have $\class{E}_{(\sfn, \sfd)}(g) \geq 0$ for all classifiers $g : \class{X} \times [K] \to \{0, 1\}$.
\end{mylemma}
% \begin{lemma}
% \label{lem:excess_score}
% Let Assumption~\ref{ass:continuity0} be satisfied.
% Let $g^*_{(\sfn, \sfd)}$ be defined as in Theorem~\ref{thm:fair_optimal_LF} and assume that $\theta^* \triangleq \theta^*_{(\sfn, \sfd)}$ defined in Eq.~\eqref{eq:fixed_point_LF} exists.
% Let $\bar{\mu}(\eta)$ be the Wasserstein barycenter of measures $\mu_1(\eta), \ldots, \mu_K(\eta)$ weighted by $p_1, \ldots, p_K$, respectively. Define $\beta^*$ as $\beta^* = F_{\bar{\mu}(\eta)}(\theta^*)$.
% Let $\class{E}_{(\sfn, \sfd)}(g) \triangleq {\rm U}_{(\sfn, \sfd)}\big(g^*_{(\sfn, \sfd)}\big) - {\rm U}_{(\sfn, \sfd)}\big(g\big)$.
% Then, for any classifier $g \in \dom({\rm U}_{(\sfn, \sfd)})$ such that $g(\bX, S) \independent S$, it holds that
% \begin{equation*}
%      \class{E}_{(\sfn, \sfd)}(g) =
%      \begin{small}
%      \frac{\Exp|{\eta(\bX, S) {-} F_{\mu_S(\eta)}^{-1}(\beta^*)}|\ind{g^*(\bX, S) {\neq} g(\bX, S)}}{{\sfd_0 {+} \sfd_1 \Prob(Y {=} 1,\, g(\bX, S) {=} 1) {+} \sfd_2\Prob(g(\bX, S) {=} 1)}}\cdot
%      \begin{dcases}
%      \frac{\sfd_2\sfn_1 {-} \sfn_2\sfd_1}{\sfd_2 + \theta^*\sfd_1} & \sfn_2\sfd_1 {\neq} \sfd_2\sfn_1\\
%      \frac{\sfn_1\sfd_0 - \sfd_1\sfn_0}{\sfd_0 {+} \sfd_1\Exp(f^*(\bX, S) {-} \theta^*)_+} &\sfn_2\sfd_1 {=} \sfd_2\sfn_1
%      \end{dcases}\enspace.
%      \end{small}
% \end{equation*}
% Furthermore, under the conditions on $(\sfn, \sfd)$ specified in Theorem~\ref{thm:fair_optimal_LF}; we have $\class{E}_{(\sfn, \sfd)}(g) \geq 0$ for all classifiers $g : \class{X} \times [K] \to \{0, 1\}$.
% \end{lemma}
%\evg{Remark is new}
\begin{remark}
\label{rem:positive_excess}
Lemma \ref{lem:excess_score}, together with  Lemma~\ref{lem:relating_bad_boy_to_bad_number}, stated in appendix, implies that
\begin{align*}
    \class{E}_{(\sfn, \sfd)}(g) =
     \parent{\sfn_1 - \sfd_1{\rm U}_{(\sfn, \sfd)}(g^*_{(\sfn, \sfd)})}\frac{\Exp|{\eta(\bX, S) {-} F_{\mu_S(\eta)}^{-1}(\beta^*)}|\ind{g^*(\bX, S) {\neq} g(\bX, S)}}{{\sfd_0 {+} \sfd_1 \Prob(Y {=} 1,\, g(\bX, S) {=} 1) {+} \sfd_2\Prob(g(\bX, S) {=} 1)}}\enspace.
\end{align*}
Hence, the inequality $\class{E}_{(\sfn, \sfd)}(g) \geq 0$ for all $g$ is implied from
\begin{align*}
    \begin{dcases}
        \sfd_0 + \sfd_1 \Prob(Y = 1,\, g(\bX, S) = 1) + \sfd_2\Prob(g(\bX, S) = 1) > 0 \qquad \forall g \in \dom({\rm U}_{(\sfn, \sfd)})\\
        \sfn_1 - \sfd_1{\rm U}_{(\sfn, \sfd)}(g^*_{(\sfn, \sfd)}) \geq 0 
    \end{dcases}\enspace.
\end{align*}
The first of the above conditions is ensured if $\sfd_0 + \min \big\{\min\{\sfd_1,\, 0\} + \sfd_2,\, 0\big\} \geq 0$ (Lemma~\ref{lem:denominator_positive}) assumed in Theorem~\ref{thm:fair_optimal_LF} and the second one is ensured by~\eqref{eq:condition1} or~\eqref{eq:condition2}, as proved in Lemma \ref{lem:sign_of_a_bad_boy}.
\end{remark}
\begin{proof}[Proof of Lemma~\ref{lem:excess_score}]
Let $\bar{\mu}(\eta)$ be the Wasserstein barycenter of measures $\mu_1(\eta), \ldots, \mu_K(\eta)$, weighted by $p_1, \ldots, p_K$ respectively. 
Assumption~\ref{ass:continuity0} and the form of $f^*$ ensures that the fair optimal classifier in Eq.~\eqref{eq:fair_optimal_LF} can be expressed as
\begin{align*}
    g^*(\bx, s) = \ind{\eta(\bx, s) \geq F_{\mu_s(\eta)}^{-1} \circ F_{\bar{\mu}(\eta)}\big(\theta^* \big)} = \ind{\eta(\bx, s) \geq F_{\mu_s(\eta)}^{-1}(\beta^*)}\enspace,
\end{align*}
where $\beta^* = F_{\bar{\mu}(\eta)}(\theta^*)$.
Fix an arbitrary classifier $g$ which satisfies the demographic parity constraint.\\
Our goal is to develop ${\rm U}(g^*) - {\rm U}(g)$, which we express as a sum of two terms $\sf{I} + \sf{II}$, with
\begin{align*}
    {\sf I} \triangleq \frac{\sfn_1\parent{\Exp[\eta(\bX, S)(g^*(\bX, S) - g(\bX, S))]} + \sfn_2\Exp[g^*(\bX, S) - g(\bX, S)]}{\sfd_0 + \sfd_1\Exp[\eta(\bX, S)g^*(\bX, S)] + \sfd_2\Exp[g^*(\bX, S)]}\enspace,
\end{align*}
and
\begin{align*}
    {\sf II} \triangleq -{\rm U}(g)\frac{\sfd_1\parent{\Exp[\eta(\bX, S)(g^*(\bX, S) - g(\bX, S))]} + \sfd_2\Exp[g^*(\bX, S) - g(\bX, S)]}{\sfd_0 + \sfd_1\Exp[\eta(\bX, S)g^*(\bX, S)] + \sfd_2\Exp[g^*(\bX, S)]}\enspace.
\end{align*}
One verifies that indeed ${\rm U}(g^*) - {\rm U}(g) = {\sf I} + {\sf II}$.
Thanks to the alternative definition of $g^*$ introduced in the beginning of this proof, for any $a, b \in \bbR$ we have
\begin{align*}
    a\Exp[\eta(\bX, S)(g^*(\bX, S) - g(\bX, S))] &+ b\Exp[g^*(\bX, S) - g(\bX, S)]\\
    &= a\Exp\left[\abs{\eta(\bX, S) - F_{\mu_S(\eta)}^{-1}(\beta^*)}\ind{g^*(\bX, S) \neq g(\bX, S)}\right] \\
    &\phantom{+} +
    \Exp[(b + aF_{\mu_S(\eta)}^{-1}(\beta^*))(g^*(\bX, S)) - g(\bX, S)]\\
    &=
    a \Exp\left[\abs{\eta(\bX, S) - F_{\mu_S(\eta)}^{-1}(\beta^*)}\ind{g^*(\bX, S) \neq g(\bX, S)}\right]  \\
    &\phantom{+} +
    (b + aF_{\bar{\mu}(\eta)}^{-1}(\beta^*))\Exp[g^*(\bX, S)) - g(\bX, S)]\enspace,
\end{align*}
where the last equality is due to the fact that $g$ satisfies the demographic parity constraint.
Thus, setting $\Delta(g^*, g) \triangleq \Exp\left[|{\eta(\bX, S) - F_{\mu_S(\eta)}^{-1}(\beta^*)}|\ind{g^*(\bX, S) \neq g(\bX, S)}\right]$ and recalling that $\theta^* = F^{-1}_{\bar{\mu}(\eta)}(\beta^*)$ we can express ${\sf I}$ and ${\sf II}$ as
\begin{align*}
    &{\sf I} = \frac{\sfn_1\Delta(g^*, g) + (\sfn_2 + \sfn_1 \theta^*)\Exp[g^*(\bX, S) - g(\bX, S)]}{\sfd_0 + \sfd_1\Exp[\eta(\bX, S)g^*(\bX, S)] + \sfd_2\Exp[g^*(\bX, S)]}\enspace,\\
    &{\sf II} = -{\rm U}(g)\frac{\sfd_1\Delta(g^*, g) + (\sfd_2 + \sfd_1 \theta^*)\Exp[g^*(\bX, S) - g(\bX, S)]}{\sfd_0 + \sfd_1\Exp[\eta(\bX, S)g^*(\bX, S)] + \sfd_2\Exp[g^*(\bX, S)]}\enspace.
\end{align*}

{\textbf Case 1: $\sfn_2\sfd_1 \neq \sfd_2\sfn_1$}.
Lemma~\ref{lem:fixed_at_optim_LF} implies that
\begin{align*}
    &{\sf I} = \frac{\sfn_1\Delta(g^*, g) + {\rm U}(g^*)(\sfd_2 + \sfd_1 \theta^*)\Exp[g^*(\bX, S) - g(\bX, S)]}{\sfd_0 + \sfd_1\Exp[\eta(\bX, S)g^*(\bX, S)] + \sfd_2\Exp[g^*(\bX, S)]}\enspace.
\end{align*}
Combining the above two expressions for ${\sf I}$ and ${\sf II}$ we obtain
\begin{align*}
    {\rm U}(g^*) - {\rm U}(g)
    &=
    \parent{{\rm U}(g^*) - {\rm U}(g) } \frac{(\sfd_2 + \sfd_1 \theta^*)\Exp[g^*(\bX, S) - g(\bX, S)]}{\sfd_0 + \sfd_1\Exp[\eta(\bX, S)g^*(\bX, S)] + \sfd_2\Exp[g^*(\bX, S)]}\\
    &\phantom{=}
    +\parent{\sfn_1 - {\rm U}(g)\sfd_1}\frac{\Delta(g^*, g)}{\sfd_0 + \sfd_1\Exp[\eta(\bX, S)g^*(\bX, S)] + \sfd_2\Exp[g^*(\bX, S)]}\enspace.
\end{align*}
Simplifying the above and using Lemma~\ref{lem:f_star_eta_replacing}, we obtain
\begin{align*}
    {\rm U}(g^*) - {\rm U}(g)
    &=
    \parent{\sfn_1 - {\rm U}(g)\sfd_1}\frac{\Delta(g^*, g)}{\sfd_0 + \sfd_1\Exp\left[(f^*(\bX, S) - \theta^*)_+\right]  + (\sfd_2 + \theta^*\sfd_1)\Exp[g(\bX, S)]}\enspace.
\end{align*}
As in Lemma~\ref{lem:fixed_at_optim_LF} (using the expression for the numerator), we deduce that
\begin{align*}
    \sfd_0 + \sfd_1\Exp\left[(f^*(\bX, S) - \theta^*)_+\right]  &+ (\sfd_2 + \theta^*\sfd_1)\Exp[g(\bX, S)] \\
    &= \frac{(\sfd_2 + \theta^* \sfd_1)\parent{(\sfn_0\sfd_1 - \sfd_0\sfn_1) + (\sfn_2\sfd_1 - \sfd_2\sfn_1) \Exp[g(\bX, S)]}}{\sfn_2\sfd_1 - \sfd_2\sfn_1}\enspace,
\end{align*}
and using the definition of ${\rm U}(g)$, we can write
\begin{align}
    \label{eq:excess_LF1}
    \sfn_1 - {\rm U}(g)\sfd_1 = \frac{(\sfn_1\sfd_0 - \sfd_1\sfn_0) + (\sfn_1\sfd_2 - \sfd_1 \sfn_2)\Exp[g(\bX, S)]}{\sfd_0 + \sfd_1 \Exp[\eta(\bX, S)g(\bX, S)] + \sfd_2\Exp[g(\bX, S)]}\enspace.
\end{align}
Combining the last three displays, we arrive at the claimed equality
\begin{align*}
    {\rm U}(g^*) - {\rm U}(g) = \frac{\sfd_2\sfn_1 - \sfn_2\sfd_1}{\sfd_2 + \theta^*\sfd_1}\cdot\frac{\Exp|{\eta(\bX, S) - F_{\mu_S(\eta)}^{-1}(\beta^*)}|\ind{g^*(\bX, S) \neq g(\bX, S)}}{{\sfd_0 + \sfd_1 \Exp[\eta(\bX, S)g(\bX, S)] + \sfd_2\Exp[g(\bX, S)]}}\enspace.
\end{align*}
{\textbf Case 2: $\sfn_2\sfd_1 = \sfd_2\sfn_1$}. We have shown in the proof of Lemma~\ref{lem:fixed_at_optim_LF} that in this particular case, $\sfn_1 \theta^* + \sfn_2 = \sfd_1 \theta^* + \sfd_2 = 0$. Hence ${\sf I}$ and ${\sf II}$ reduce to
\begin{align*}
    &{\sf I} = \frac{\sfn_1\Delta(g^*, g)}{\sfd_0 + \sfd_1\Exp[\eta(\bX, S)g^*(\bX, S)] + \sfd_2\Exp[g^*(\bX, S)]}\enspace,\\
    &{\sf II} = -{\rm U}(g)\frac{\sfd_1\Delta(g^*, g)}{\sfd_0 + \sfd_1\Exp[\eta(\bX, S)g^*(\bX, S)] + \sfd_2\Exp[g^*(\bX, S)]}\enspace.
\end{align*}
Consequently, the difference of utilities is expressed as
\begin{align*}
    {\rm U}(g^*) - {\rm U}(g)
    &=
    \parent{\sfn_1 - {\rm U}(g)\sfd_1}\frac{\Delta(g^*, g)}{\sfd_0 + \sfd_1\Exp[\eta(\bX, S)g^*(\bX, S)] + \sfd_2\Exp[g^*(\bX, S)]}\enspace.
\end{align*}
Again invoking the result of Lemma~\ref{lem:f_star_eta_replacing}, we deduce
%Using Lemma~\ref{lem:fixed_at_optim_LF}, 
\begin{align*}
\sfd_0 + \sfd_1\Exp[\eta(\bX, S)g^*(\bX, S)] + \sfd_2\Exp[g^*(\bX, S)] &= \sfd_0 + \sfd_1\Exp\left[(f^*(\bX, S) - \theta^*)_+\right]\enspace.
% &= \frac{\sfn_0 + \sfn_1\Exp\left[(f^*(\bX, S) - \theta^*)_+\right]}{{\rm U}(g^*)}\enspace.
\end{align*}
The above two displays combined with Eq.~\eqref{eq:excess_LF1} and the condition $\sfn_2\sfd_1 = \sfd_2\sfn_1$ yield
\begin{align*}
    {\rm U}(g^*) - {\rm U}(g)
    &=
    \frac{\sfn_1\sfd_0 - \sfd_1\sfn_0}{\sfd_0 + \sfd_1\Exp\left[(f^*(\bX, S) - \theta^*)_+\right]}\cdot\frac{\Delta(g^*, g)}
    {\sfd_0 + \sfd_1 \Exp[\eta(\bX, S)g(\bX, S)] + \sfd_2\Exp[g(\bX, S)]}\enspace.
\end{align*}
The proof is concluded.
\end{proof}
Let us remark that the content of this section can be seen as a strict improvement over~\cite{koyejo2014consistent} who only derived Lemma~\ref{lem:fixed_at_optim_LF} in the absence of the fairness constraint. Indeed, assuming that $S \independent \bX$, ensures that \emph{any} classifier $g$ is demographic parity fair and that $f^* \equiv \eta$. In the absence of the demographic parity constraint, Assumption~\ref{ass:continuity0} is not necessary and \emph{exactly} the same proof technique allows to obtain the characterization of the optimal unconstrained classifier.

\paragraph{Examples: accuracy and ${\rm F}_1$-score} 
In this part, we give specific examples of the parameters $(\sfn_0, \sfn_1, \sfn_2)$ and $(\sfd_0, \sfd_1, \sfd_2)$ and instantiate Theorem~\ref{thm:fair_optimal_LF}
and Lemma~\ref{lem:excess_score}. The first examples concerns the accuracy as a performance metric. It highlights the generality of the derived results.
\begin{example}[Accuracy under fairness constraint]
Recalling the coefficients specified in Example~\ref{ex:accuracy}, we see that in this case $\sfn_2\sfd_1 - \sfd_2\sfn_1 = -1 \cdot 0 - 0 \cdot 2 = 0$. Furthermore, one checks that condition~\eqref{eq:condition2} is satisfied. Hence under Assumption~\ref{ass:continuity0},
Theorem~\ref{thm:fair_optimal_LF} states that
\begin{align*}
    g^*_{({\sf n}, {\sf d})}(\bx, s) = \ind{f^*(\bx, s) \geq \theta^*_{({\sf n}, {\sf d})}}\enspace,
\end{align*}
with $\theta^*_{({\sf n}, {\sf d})} = \tfrac{\sfd_0\sfn_2-\sfn_0\sfd_2}{\sfn_2\sfd_1 - \sfd_2\sfn_1} = \tfrac{1 \cdot (-1) - (1 - \Prob(Y = 1))\cdot 0}{(1 - \Prob(Y = 1)) \cdot 0 - 1 \cdot 2} = \tfrac{1}{2}$ maximizes $\Prob(Y \neq g(\bX, S))$ under the demographic parity constraint. Thus, it coincides with the result of Theorem~\ref{thm:equivalence}. 
Furthermore, Lemma~\ref{lem:excess_score} states that for any classifier $ g : \class{X} \times \class{S} \to \{0, 1\}$ such that $ g(\bX, S) \independent S$, it holds that
\begin{align*}
    \Prob(Y {=} g^*_{({\sf n}, {\sf d})}(\bX, S)) {-} \Prob(Y {=}  g(\bX, S)) = 2\Exp|{\eta(\bX, S) {-} F_{\mu_S(\eta)}^{-1}\circ F_{\bar{\mu}(\eta)}(.5)}|\ind{g^*(\bX, S) {\neq} g(\bX, S)}\enspace.
\end{align*}
We invite the reader to compare the above expression with its unconstrained version~\cite[Theorem 2.2]{devroye2013probabilistic}.
\end{example}

The second example concerns the ${\rm F}_1$-score that has been used in several empirical works on fairness as a performance measure~\citep{wang2021analyzing,dablain2022towards,wick2019unlocking}.
\begin{example}[${\rm F}_1$-score under fairness constraint]
Recall that the ${\rm F}_1$-score is defined as
\begin{align*}
    {\rm F}_1(g) = \frac{2\Prob(g(\bX, S) = 1,\, Y = 1)}{\Prob(Y = 1) + \Prob(g(\bX, S) = 1)}\enspace.
\end{align*}
Using the coefficients specified in Example~\ref{ex:F_score}, we see that for this case $\sfn_2\sfd_1 \neq \sfd_2\sfn_1$ and condition~\eqref{eq:condition1} is satisfied.
Hence, under Assumption~\ref{ass:continuity0},
Theorem~\ref{thm:fair_optimal_LF} states that
\begin{align*}
    g^*_{({\sf n}, {\sf d})}(\bx, s) = \ind{f^*(\bx, s) \geq \theta^*_{({\sf n}, {\sf d})}}\enspace,
\end{align*}
with $\theta^*_{({\sf n}, {\sf d})}$ being a unique solution of
\begin{align*}
    \Prob(Y = 1)\theta = \Exp(f^*(\bX, S) - \theta)_+\enspace,
\end{align*}
maximizes the ${\rm F}_1$-score under the demographic parity constraint. Furthermore, Lemma~\ref{lem:excess_score} states that for any classifier $ g : \class{X} \times \class{S} \to \{0, 1\}$ such that $ g(\bX, S) \independent S$, it holds that
\begin{align*}
    {\rm F}_1\big(g^*_{({\sf n}, {\sf d})}\big) - {\rm F}_1\big( g\big) = \frac{2\Exp\big|{\eta(\bX, S) - F_{\mu_S(\eta)}^{-1}\circ F_{\bar{\mu}(\eta)}\big(\theta^*_{({\sf n}, {\sf d})}\big)}\big|\ind{g^*_{({\sf n}, {\sf d})}(\bX, S) \neq {g}(\bX, S)}}{\Prob(Y = 1) \,+\, \Prob(g(\bX, S) = 1)}\enspace.
\end{align*}
We invite the reader to compare the above expression with its unconstrained version~\cite[Lemma 2]{chzhen2020optimal}.
\end{example}

\section{The unawareness case}
\label{sec:unwareness}
All the previous parts were concerned with the awareness setup---we allowed ourselves to use the sensitive attribute explicitly. However, it can happen in practice that for legal or ethical reasons, the sensitive attribute cannot be used as an input at prediction time \citep{barocas2016big}.
Throughout this section we look at classifiers of the form $g : \class{X} \to \{0, 1\}$. By abuse of notation, and as long as confusion cannot occur, we use the same notation $\class{G}$ to denote the set of all classifiers in the unawareness setup. We also need to introduce the conditional distribution of the sensitive attribute $S$, given the nominally non-sensitive features $\bX$. For all $s \in [K]$, we set $\tau_s(\bX) = \Prob(S = s \mid \bX)$. With one more abuse of notation, we set $\eta(\bX) \triangleq \Exp[Y \mid \bX]$.
In this section we look for
\begin{highlighted}
\begin{align}
    \label{eq:def_optinmal_unawareness}
    g^* \in \argmin_{g \in \class{G}}\enscond{\Prob(g(\bX) \neq Y)}{g(\bX) \independent S}\enspace.
\end{align}
\end{highlighted}
Note that the only difference with the previous setup is the absence of the sensitive input $S$ in the input of $g$. \citet{lipton2018does} investigated this framework empirically and provided evidence against its use in practice. In particular, they empirically showed that while not permitting using the sensitive attribute $S$, many algorithms still learn the link between $S$ and $\bX$ implicitly. Our first result gives a theoretical justification to this phenomenon.

As in the awareness case, we work under a continuity assumption, adapted to this scenario. Recall that Assumption~\ref{ass:continuity0} imposed continuity of the regression function distribution $\Law(\eta(\bX,s))$ for each sensitive group $s \in S$. Here we need a different assumption to account for the fact that $S$ is not accessible anymore, namely the continuity of any linear combination of the regression functions distributions $\eta(\bX)$ and $(\tau_s(\bX))_{s \in K}$.

\begin{assumption}
\label{ass:continuity_unawareness0}
    For every $s \in [K]$ and for every vector $\bc = (c_1, \ldots, c_K)^\top \in \mathbb{R}^{K}$ such that $c_1 + \ldots + c_K = 0$, the distribution $\Law(\eta(\bX) + \sum_{\sigma=1}^K \tfrac{c_\sigma}{p_{\sigma}} \tau_\sigma(\bX) \mid S = s)$ is continuous.
\end{assumption}

Akin to Theorem~\ref{thm:optimal_DP_0}, we derive the explicit form of an optimal fair classifier in the unawareness setting.
\begin{theorem}
\label{thm:optimal_DP_unawareness0}
Let Assumption~\ref{ass:continuity_unawareness0} be satisfied. Then a solution $g^*$ defined in Eq.~\eqref{eq:def_optinmal_unawareness} can be expressed for all $\bx \in \class{X}$ as
\begin{align*}
     &g^*(\bx) = \ind{2\eta(\bx) - 1 \geq \sum_{\sigma =1}^K \frac{\lambda_{\sigma }^* \tau_{\sigma }(\bx)}{p_{\sigma }}}\enspace,
\end{align*}
where $\blambda^* = (\lambda_1^*, \ldots, \lambda_K^*) \in \bbR^K$ is a solution of
\begin{align}
    \label{eq:lambda_dp_awareness}
     \min_{\blambda \in \bbR^K} \enscond{\Exp\parentsq{\abs{ 2\eta(\bX) - 1 - \sum_{\sigma =1}^K \frac{\lambda_{\sigma }\tau_{\sigma }(\bX)}{p_{\sigma }}}}}{\Exp\left[\frac{\lambda_S}{p_S}\right] = 0}\enspace.
\end{align}
\end{theorem}
We make two observations. First of all, the optimal fair classifier is no longer given by the group-wise threshold. Yet, one can think of the term $\theta(\bx) \triangleq \sum_{\sigma =1}^K \frac{\lambda_{\sigma }^* \tau_{\sigma }(\bx)}{p_{\sigma }}$ as the $\bx$-dependent threshold. The optimal classifier $g^*$ tries to guess the value of the sensitive attribute from the features to properly set the threshold. Note that as in the awareness case, here we have $\Exp[\theta(\bX)] = 0$. Thus, in average, the ``threshold'' remains being equal to $1/2$ as in the standard classification setup. Secondly, we see that if $S$ is measurable w.r.t. $\bX$, we fall back to the awareness case. Otherwise each variable $\lambda^*_s$ is weighted by the conditional distribution of $S$ given $\bX$.

Importantly, it is remains an open problem to give a connection of the above problem with the corresponding regression setup. The main reason for it is the current lack of an explicit solution to the optimal fair regression problem in the unawareness case. Some attempts were made in~\citep{chzhen2020example}, yet they are unsatisfactory and do not give a complete picture. Intuitively, the difficulty of extending the optimal transport based approach to the unawareness setup lies in our inability to establish the source of a given $\bx$. In other words, given $\bx$, we have no idea which of $\Prob_{\bX \mid S = 1}, \ldots, \Prob_{\bX \mid S = K}$ it was sampled from. Hence, we cannot build a transport map from $\Law(\eta(\bX, S) \mid S = s)$ to their common barycenter since it requires the knowledge of $S$. Naively, one might think to use $\hat{S}(\bX)$---the best prediction of $S$ given $\bX$---instead of $S$. While intuitive, it is easy to see that simply replacing $S$ by $\hat{S}(\bX)$ in Theorem~\ref{thm:equivalence} does not even satisfy the demographic parity constraint in general. As we show in the next paragraph, the connection between the fair classification and fair regression can be made explicit in the unawareness case if we consider the case of $K = 2$. The existence of such a connection is explained by the Hahn decomposition theorem for signed measure, whose generalization (even its formulation) to many measures is unclear.

\paragraph{Binary sensitive attribute: the $(\Prob \to \QProb)$ reduction.}
In this section we describe a reduction of the fair unaware binary classification problem to the awareness case for $K = 2$. First of all, let us recall that the minimization of $\Prob(Y \neq g(\bX, S))$ over $g$ under any constraints is equivalent to the minimization of $\Exp[g(\bX, S)(1 - 2\eta(\bX, S))]$ under the same constraints. Furthermore, the same applies to the awareness case where we only need to replace $\eta(\bX, S)$ by $\eta(\bX)$.

For our reduction, given a distribution $\Prob$ on $\class{X} \times \{1 , 2\} \times \{0, 1\}$, we build \emph{another} distribution $\QProb$ on $\class{X} \times \{1, 2\}$ and a function $\tilde\eta: \class{X} \times \{1, 2\} \to [0, +\infty)$ with the following property: there is a one-to-one correspondence between
\begin{align*}
    g^*_{\Prob} \in \argmin_{g: \class{X}\to \{0, 1\}}\enscond{\Exp_{\Prob}[g(\bX)(1 - 2\eta(\bX))]}{g(\bX) \independent_{\Prob} S}\enspace,
\end{align*}
and
\begin{align*}
    g^*_{\QProb} \in \argmin_{g: \class{X} \times \{1, 2\}\to \{0, 1\}}\enscond{\Exp_{\QProb}[g(\bX, S)(1 - 2\tilde{\eta}(\bX, S))]}{g(\bX, S) \independent_{\QProb} S}\enspace.
\end{align*}
In other words, if $g^*_{\QProb}$ is an optimal fair classifier for distribution $\QProb$ under \emph{awareness}, then $g^*_{\QProb}$ can be transformed into an optimal fair classifier $g^*_{\Prob}$ for $\Prob$ under \emph{unawareness}.
In what follows, we present the reduction and, given the distribution $\Prob$, explain the procedure to build $\QProb$.

Let $\TV \triangleq \tfrac{1}{2}\int \abs{\d \Prob_{\bX \mid S = 1} - \d \Prob_{\bX \mid S = 2}}$. Note that if $\TV = 0$, then $\bX \independent S$ and any unaware classifier satisfies the demographic parity constraint. Hence, we assume that $\TV \in (0, 1]$. We define $\QProb$ in three steps.
\begin{highlighted}
\begin{enumerate}
    \item[\bf Step 1.] The distribution of $\bX$ given $S$ under $\QProb$ is defined as
    \begin{align*}
    \QProb_{\bX \mid S = 1} = \frac{(\Prob_{\bX \mid S = 1} - \Prob_{\bX \mid S = 2})_+}{\TV}\qquad\text{and}\qquad\QProb_{\bX \mid S = 2} = \frac{(\Prob_{\bX \mid S = 2} - \Prob_{\bX \mid S = 1})_+}{\TV}\enspace,
    \end{align*}
    where $(\Prob_{\bX \mid S = 1} - \Prob_{\bX \mid S = 2})_+$ and $(\Prob_{\bX \mid S = 2} - \Prob_{\bX \mid S = 1})_+$ is the Hahn decomposition of the signed measure $\Prob_{\bX \mid S = 2} - \Prob_{\bX \mid S = 1}$~\citep[see, e.g.,][Theorem 32.1]{billingsley2008probability};
    % \item[\bf Step 2.] the marginal of $S$ under $\QProb$ is defined as $\QProb(S = 1) = \QProb(S = 2) = \tfrac{1}{2}$;
    \item[\bf Step 2.] the distribution of $S$ under $\QProb$ is defined as
    \begin{align*}
    \QProb(S = 1) = \QProb(S = 2) = \frac{1}{2}\enspace;
    \end{align*}
    \item[\bf Step 3.] the new pseudo-regression function $\tilde{\eta}$ is defined as
\begin{align*}
    \tilde{\eta}(\bx, s) = \frac{1}{2} + \frac{\TV}{2} \cdot \frac{2\eta(\bx) - 1}{\abs{\frac{\tau_1(\bx)}{p_1} - \frac{\tau_2(\bx)}{p_2}}}\quad \text{for}\quad \bx \in \supp(\QProb_{\bX \mid S = 1}) \cap \supp(\QProb_{\bX \mid S = 2})\enspace;
\end{align*}
\end{enumerate}
\end{highlighted}
We note that under $\QProb$, the sensitive attribute $S$ is measurable w.r.t. $\bX$ since the supports of $\QProb_{\bX \mid S = 1}$ and $\QProb_{\bX \mid S = 2}$ do not intersect. We refer $\tilde{\eta}$ as to the pseudo-regression function since it is not guaranteed that it takes values in $[0, 1]$ and, hence, is not necessary a valid regression function of $Y \mid \bX$ under $\QProb$ for $Y \in \{0, 1\}$.
\begin{myproposition}{Unawareness to awareness reduction}{}
    Let $\Prob$ be any distribution on $\class{X} \times \{1, 2\} \times \{0, 1\}$. Let $\QProb$ and $\tilde{\eta}$ be defined using the three steps procedure described above and
    \begin{align*}
        g^*_{\QProb} \in \argmin_{g : \class{X} \times \{1, 2\} \to \{0, 1\}}\enscond{\Exp_{\QProb}[g(\bX, S)(1 - 2\tilde{\eta}(\bX, S))]}{g(\bX, S) \independent_{\QProb} S}\enspace.
    \end{align*}
Then, $g^*_{\Prob}: \class{X} \to \{0, 1\}$ defined point-wise as
\begin{align*}
    g^*_{\Prob}(\bx) =
    \begin{cases}
    g^*_{\QProb}(\bx, 1) & \bx \in \supp(\QProb_{\bX \mid S = 1})\\
    g^*_{\QProb}(\bx, 2) & \bx \in \supp(\QProb_{\bX \mid S = 2})\\
    \ind{\eta(\bx) \geq 1/2} & \bx \in \supp(\Prob_{\bX}) \setminus \parent{\supp(\QProb_{\bX \mid S = 1}) \cup \supp(\QProb_{\bX \mid S = 2})}
    \end{cases}\enspace,
\end{align*}
is a solution of $\min_{g : \class{X} \to \{0, 1\}}\enscond{\Exp_{\Prob}[g(\bX)(1 - 2\eta(\bX))]}{g(\bX) \independent_{\Prob} S}$.
\end{myproposition}
% \begin{proposition}
%     Let $\Prob$ be any distribution on $\class{X} \times \{1, 2\} \times \{0, 1\}$. Let $\QProb$ and $\tilde{\eta}$ be defined using the three steps procedure described above and
%     \begin{align*}
%         g^*_{\QProb} \in \argmin_{g : \class{X} \times \{1, 2\} \to \{0, 1\}}\enscond{\Exp_{\QProb}[g(\bX, S)(1 - 2\tilde{\eta}(\bX, S))]}{g(\bX, S) \independent_{\QProb} S}\enspace.
%     \end{align*}
% Then, $g^*_{\Prob}: \class{X} \to \{0, 1\}$ defined point-wise as
% \begin{align*}
%     g^*_{\Prob}(\bx) =
%     \begin{cases}
%     g^*_{\QProb}(\bx, 1) & \bx \in \supp(\QProb_{\bX \mid S = 1})\\
%     g^*_{\QProb}(\bx, 2) & \bx \in \supp(\QProb_{\bX \mid S = 2})\\
%     \ind{\eta(\bx) \geq 1/2} & \bx \in \supp(\Prob_{\bX}) \setminus \parent{\supp(\QProb_{\bX \mid S = 1}) \cup \supp(\QProb_{\bX \mid S = 2})}
%     \end{cases}\enspace,
% \end{align*}
% is a solution of $\min_{g : \class{X} \to \{0, 1\}}\enscond{\Exp_{\Prob}[g(\bX)(1 - 2\eta(\bX))]}{g(\bX) \independent_{\Prob} S}$.
% \end{proposition}
\begin{proof}
For any $g : \class{X} \times \{1, 2\} \to \{0, 1\}$, define $\tilde{g} : \class{X} \to \{0, 1\}$ as
\begin{align*}
    \tilde{g}(\bx) = 
    \begin{cases}
    g(\bx, 1) & \bx \in \supp(\QProb_{\bX \mid S = 1})\\
    g(\bx, 2) & \bx \in \supp(\QProb_{\bX \mid S = 2})\\
    \ind{\eta(\bx) \geq 1/2} & \bx \in \supp(\Prob_{\bX}) \setminus \parent{\supp(\QProb_{\bX \mid S = 1}) \cup \supp(\QProb_{\bX \mid S = 2})}
    \end{cases}\enspace.
\end{align*}
Note that the above correspondence of $g$ and $\tilde{g}$ is invertible since the supports of $\QProb_{\bX \mid S = 1}$ and $\QProb_{\bX \mid S = 2}$ do not intersect by construction.
Observe that for any $g : \class{X} \times \{1, 2\} \to \{0, 1\}$ it holds that
\begin{align*}
    g(\bX, S) \independent_{\QProb} S \,\Longleftrightarrow\,
    g(\cdot, 1) \sharp \QProb_{\bX \mid S = 1} = g(\cdot, 2) \sharp \QProb_{\bX \mid S = 2}
    \,\Longleftrightarrow\,
    \tilde{g} \sharp \Prob_{\bX \mid S = 1} = \tilde{g} \sharp \Prob_{\bX \mid S = 2}\enspace.
\end{align*}
Thus, given any classifier $g$ satisfying the demographic parity constraint under $\QProb$, we can transform it to a classifier that satisfies the constraints under $\Prob$.
Furthermore, since
\begin{align*}
     \Exp_{\QProb}[g(\bX, S)(1 - 2\tilde{\eta}(\bX, S))]
     =
     \Exp_{\Prob}[\tilde g(\bX)(1 - 2Y)\ind{\bX \in \supp(\QProb_{\bX \mid S = 1}) \cap \supp(\QProb_{\bX \mid S = 2})}]\enspace,
\end{align*}
taking any classifier $\bar{g} : \class{X} \to \{0, 1\}$ we can write
\begin{align*}
    \Exp_{\Prob}[\bar g(\bX)(1 - 2Y)]
    &=
    \Exp_{\QProb}[\bar g(\bX, S)(1 - 2\tilde{\eta}(\bX, S))\ind{\bX \in \supp(\QProb_{\bX \mid S = 1}) \cap \supp(\QProb_{\bX \mid S = 2})}]\\
    &\phantom{=}
    + \Exp_{\Prob}[\bar g(\bX)(1 - 2Y)\ind{\bX \not\in \supp(\QProb_{\bX \mid S = 1}) \cap \supp(\QProb_{\bX \mid S = 2})}]\enspace,
\end{align*}
where in the first equality, we added the input $S$ to $\bar{g}$ sue to the fact that $S$ is $\bX$ measurable under $\QProb$. Note that the second term is minimized point-wise by the Bayes classifier, while the first term is minimized by $g^*_{\QProb}$ thanks to the equivalence established for the demographic parity constraint.
\end{proof}

The above result provide a theoretical justification to the empirical observations made by~\cite{lipton2018does}. Indeed, they have empirically shown that in the unawareness setting, many classification algorithms tailored for the demographic parity constraint, are forced to ``guess'' the sensitive attribute $S$. Theoretically, this is reflected by the construction of the distribution $\bX \mid S$ under $\QProb$. Furthermore, since the reduction is performed to the awareness setup, the results of previous sections on the connection between fair regression and fair classification still applies. Yet, we emphasize that the above argument is only valid for $K = 2$ and its extension to $K > 2$ remains an open problem. The main difficulty comes from the absence of a version of the Hahn decomposition for more than two measures.

\section{Fair learning: from infinite to finite sample}
%\evg{this is new}
All the previous sections were concerned with the ``infinite sample'' regime---the case of known distribution $\Prob$. While not being the main focus of the paper, given the established connection with the problem of fair regression, one can easily pass from the infinite to the finite-sample regime.
Indeed, there are many algorithms that allow to consistently estimate the regression function $f^*$. For instance,~\cite{agarwal2019fair} give an in-processing algorithm with provable finite sample generalization bounds;~\cite{gouic2020price} propose a consistent estimator of $f^*$;~\cite{chzhen2020fair} provide an algorithm with finite sample fairness and risk guarantees;~\cite{chzhen2020minimax} exhibit a modification of the two aforementioned estimators that enjoys stronger fairness and risk guarantees.

Once an estimator $\hat{f}$ of $f^*$ is constructed, one only needs to estimate the threshold $\theta^*$ specified in Theorem~\ref{thm:fair_optimal_LF}. Recall that there are two cases considered in Theorem~\ref{thm:fair_optimal_LF}, the first one requires finding a root of a specific function and the second one gives an explicit expression for $\theta^*$. For the first case one can use the \emph{unsupervised} approach recycling $\hat f$ and only estimating $\Exp_{\bX \mid S}[\cdot]$ and the, potentially distribution dependent coefficients, $(\sfn_0, \sfn_1, \sfn_2), (\sfd_0, \sfd_1, \sfd_2)$. For the second case one only needs to estimate or substitute the values of $(\sfn_0, \sfn_1, \sfn_2), (\sfd_0, \sfd_1, \sfd_2)$. Such an approach was analyzed in~\cite{chzhen2020optimal} in the context of binary classification with the ${\rm F}_1$-score without fairness considerations. Alternatively, for the threshold estimation, one can deploy the grid-search technique proposed by~\cite{koyejo2014consistent} by again recycling the base estimator $\hat{f}$ of $f^*$. In either case one ends up with a flexible and rather direct approach for building data-driven algorithms. We note however that the second approach requires additional labeled data, while the first one is only based on the unlabeled data. The final classification algorithm eventually takes the form of $\mathbbm{1}(\hat{f}(\bx, s) \geq \hat{\theta})$.
\section{Conclusion}

%\evg{this is new}
We have derived an explicit connection between the regression and classification under the demographic parity constraint problems. Leveraging the optimal transport interpretation of the optimal fair regressor, we have shown that the regression-classification link is akin to the classical unconstrained setup. This connection is extended to non-decomposable performance measures and, remarkably, amounts to replacing the standard regression function by its fair counterpart.
Finally, we have provided a reduction scheme to pass from the unawareness setup to the awareness setup in the case of the binary sensitive attribute, hence giving the first explicit solution of the fair optimal unaware classifier. Our results are instructive and, relying on the previous studies, lead to wide spectrum of algorithms that can be used with non-decomposable measures. Future works will be focused on further clarification of other notions of fairness constraint by providing clean and interpretable theoretical studies.

\newpage
\appendix

\section{A unified proof for deriving optimal fair classifiers}
\label{sec:unified_proof}

In this section we state and prove a general result which implies both Theorem~\ref{thm:optimal_DP_0} and Theorem~\ref{thm:optimal_DP_unawareness0}.
On top of the problem setup presented in Section~\ref{sec:problem_setup}, let $\bW$ be a random variable taking its values in some abstract space $\mathcal{W}$. Moreover, define the regression functions $\tau_s(\bw) \coloneqq \Prob\left(S=s \mid \bW = \bw \right), s \in [K]$. The random variable $\bW$ should be thought as  $(\bX, S)$ for the awareness setting and $\bX$ for the unawareness setting. Our goal is to find a solution
\begin{highlighted}
\begin{align}
    \label{eq:def_optinmal_unified}
    g^* \in \argmin_{g \in \class{G}}\enscond{\Prob(g(\bW) \neq Y)}{g(\bW) \independent S}\enspace.
\end{align}
\end{highlighted}
The general result will be stated under the following continuity assumption. It requires continuity of the distribution of any linear combination of the regression functions evaluated at $\bW$.
\begin{assumption}
\label{ass:continuity_unified}
    For every $s \in [K]$ and for every vector $\bc = (c_1, \ldots, c_K)^\top \in \mathbb{R}^{K}$ such that $c_1 + \ldots + c_K = 0$, the distribution $\Law(\eta(\bW) + \sum_{\sigma=1}^K \tfrac{c_\sigma}{p_{\sigma}} \tau_\sigma(\bW) \mid S = s)$ is continuous.
\end{assumption}
Akin to Assumptions~\ref{ass:continuity0} and \ref{ass:continuity_unawareness0}, Assumption~\ref{ass:continuity_unified} is not necessary to prove our result but it greatly simplifies its presentation and interpretation.
Let us now state the general result which encompasses the two special cases presented in the main body of the paper.
\begin{mytheo}{Fair optimal classifier (unified version)}{optimal_DP_unified}
Let Assumption~\ref{ass:continuity_unified} be satisfied. Then a solution $g^*$ defined in Eq.~\eqref{eq:def_optinmal_unified} can be expressed for all $\bw \in \class{W}$ as
\begin{align*}
     &g^*(\bw) = \ind{2\eta(\bw) - 1 \geq \sum_{\sigma =1}^K \frac{\lambda_{\sigma }^* \tau_{\sigma }(\bw)}{p_{\sigma }}}\enspace,
\end{align*}
where $\blambda^* = (\lambda_1^*, \ldots, \lambda_K^*) \in \bbR^K$ is a solution of
\begin{align}
    \label{eq:lambda_dp_unified}
     \min_{\blambda \in \bbR^K} \enscond{\Exp\parentsq{\abs{ 2\eta(\bW) - 1 - \sum_{\sigma =1}^K \frac{\lambda_{\sigma }\tau_{\sigma }(\bW)}{p_{\sigma }}}}}{\Exp\left[\frac{\lambda_S}{p_S}\right] = 0}\enspace.
\end{align}
\end{mytheo}
% \begin{theorem}
% \label{thm:optimal_DP_unified}
% Let Assumption~\ref{ass:continuity_unified} be satisfied. Then a solution $g^*$ defined in Eq.~\eqref{eq:def_optinmal_unified} can be expressed for all $\bw \in \class{W}$ as
% \begin{align*}
%      &g^*(\bw) = \ind{2\eta(\bw) - 1 \geq \sum_{\sigma =1}^K \frac{\lambda_{\sigma }^* \tau_{\sigma }(\bw)}{p_{\sigma }}}\enspace,
% \end{align*}
% where $\blambda^* = (\lambda_1^*, \ldots, \lambda_K^*) \in \bbR^K$ is a solution of
% \begin{align}
%     \label{eq:lambda_dp_unified}
%      \min_{\blambda \in \bbR^K} \enscond{\Exp\parentsq{\abs{ 2\eta(\bW) - 1 - \sum_{\sigma =1}^K \frac{\lambda_{\sigma }\tau_{\sigma }(\bW)}{p_{\sigma }}}}}{\Exp\left[\frac{\lambda_S}{p_S}\right] = 0}\enspace.
% \end{align}
% \end{theorem}

\begin{remark}[Relating the above result to the main body]
It is straightforward to derive Theorem~\ref{thm:optimal_DP_0} and Theorem~\ref{thm:optimal_DP_unawareness0} from Theorem~\ref{thm:optimal_DP_unified}.
Indeed, to prove Theorem~\ref{thm:optimal_DP_0}, set $\bW = (\bX, S), \bw = (\bx, s)$ and notice that $\tau_\sigma(\bw) = \Prob( S = \sigma \mid \bX = \bx, S=s) = \delta_s(\sigma)$. In particular, Assumption~\ref{ass:continuity_unified} is weaker than Assumption~\ref{ass:continuity_unawareness0} and one can check that the optimal fair classifiers coincide.
Similarly, Theorem~\ref{thm:optimal_DP_unawareness0} can be derived from Theorem~\ref{thm:optimal_DP_unified} by setting $\bW = \bX, \bw = \bx$.
\end{remark}

\begin{proof}[Proof of Theorem~\ref{thm:optimal_DP_unified}]

One can verify that the minimization of $\Prob(g(\bW)) \neq Y)$ over $g$ is equivalent to the minimization of $\Exp[g(\bW)(1 - 2\eta(\bW))]$.
Furthermore, the demographic parity constraint can be equivalently expressed as
\begin{align*}
    \Exp[g(\bW) \mid S = s] = \Exp[g(\bW)]\,,\,\, s \in [K]\enspace.
\end{align*}
Thus, we are interested in the solution of the optimization problem
\begin{align*}
    \min_{g \in \class{G}}\enscond{\sum_{s \in [K]}p_s \Exp[g(\bW)(1 - 2\eta(\bW)) \mid S = s]}{\Exp[g(\bW) \mid S = s] = \Exp[g(\bW)]\,,\,\, s \in [K]}\enspace.
\end{align*}

Recall that we defined the random variable $\tau_s(\bW) = \Prob\left(S=s \mid \bW \right), s \in [K]$. The Lagrangian for the above problem can be expressed as 
\begin{align*}
     \cL(g, \blambda) = \Exp\parentsq{g(\bW)\parent{(1 - 2\eta(\bW)) - \sum_{\sigma =1}^K \lambda_{\sigma }(1- p_{\sigma }^{-1}\tau_{\sigma }(\bW)) }} \enspace,
\end{align*}
where $\blambda \in \bbR^K$. Weak duality implies that
\begin{align}
\label{eq:weak_duality}
    \min_g \max_{\blambda} \cL(g, \blambda) \geq \max_{\blambda} \min_g \cL(g, \blambda)\enspace.
\end{align}
Our approach to derive the optimal fair classifier can be decomposed in two classical steps: find optimal solutions to the dual problem $\max_{\blambda} \min_g \cL(g, \blambda)$; show that strong duality holds so that the optimal solutions to the dual problem are also optimal for the primal problem.

\paragraph{Solving the dual problem.} 
In what follows we focus our attention on the dual $\max\min$ problem, which can be solved analytically. We first solve for any $\blambda$ the inner minimization problem of the $\max\min$ formulation 
\begin{align}
    \label{eq:weak_dual}
    \min_{g}\class{L}(g, \blambda)\enspace.
\end{align}
Since $g$ can be any function from $\class{W}$ to $\{0,1\}$, the above problem can be solved point-wise. In particular, one can check that the solution is given by
\begin{align*}
    g^*(\bw) = \ind{2\eta(\bw) - 1 \geq \sum_{\sigma =1}^K \lambda_{\sigma }( p_{\sigma }^{-1}\tau_{\sigma }(\bw) - 1) }\enspace.
\end{align*}
Plugging the optimal solution $g^*$ back in the dual problem, we obtain as solution of the outer maximization problem
\begin{align}
\label{eq:lambda_opt}
    \blambda^* \in \argmin_{\blambda \in \bbR^K} \Exp\parentsq{\parent{ 2\eta(\bW) - 1 + \sum_{\sigma =1}^K \lambda_{\sigma }(1- p_{\sigma }^{-1}\tau_{\sigma }(\bW))}_+}\enspace.
\end{align}
The objective of the above optimization problem is non-negative, continuous convex as a function of $\blambda$. 
Lemma~\ref{lem:minimizer_exists} ensures that $\blambda^*$ exists.

The objective function of problem in Eq.~\eqref{eq:lambda_opt} is not smooth everywhere due to the presence of the positive part function. However, thanks to Assumption~\ref{ass:continuity_unified}, the set of points at which the objective function is not differentiable has zero Lebesgue measure and can thus be ignored~\cite[see,  e.g.,][Proposition 3]{bertsekas1973stochastic}.
The First-Order Optimality Condition (FOOC) on the optimal Lagrange multiplier $\blambda^*$ then reads as
\begin{align*}
    \Exp[p_s^{-1}\tau_s(\bW) \ind{g^*(\bW) =1}] = \Prob(g^*(\bW) =1)\,,\,\, \forall s \in [K]\enspace.
\end{align*}
The LHS of the above inequality can be simplified into
\begin{align*}
    \Exp[p_s^{-1}\tau_s(\bW) \ind{g^*(\bW) =1}] = \sum_{s=1}^K \Exp[\tau_s(\bW) \ind{g^*(\bW) =1} \mid S{=}s] = \Prob(g^*(\bW) =1 \mid S{=}s)\enspace,
\end{align*}
showing that the FOOC on $\blambda^*$ is equivalent to $g^*$ satisfying DP.

\paragraph{Strong duality.} The above reasoning showed that $g^*$ defined with the optimal Lagrange multiplier $\blambda^*$ is feasible for the primal problem. Combining this property with Eq.~\eqref{eq:weak_duality} implies that $g^*$ is also a solution of the primal problem.

\paragraph{A more convenient expression.}
Using the fact that $2(a)_+ = a + |a|$ and $\Exp\tau_s(\bW)=p_s$, we can express the optimal Lagrange multiplier $\blambda^*$ as
\begin{align*}
    \blambda^* \in \argmin_{\blambda \in \bbR^K} \Exp\parentsq{\abs{ 2\eta(\bW) - 1 + \sum_{\sigma =1}^K \lambda_{\sigma}(1- p_{\sigma }^{-1}\tau_{\sigma }(\bW))}}\enspace.
\end{align*}

Moreover, introducing $G(\blambda) = \Exp\parentsq{\abs{ 2\eta(\bW) - 1 + \sum_{\sigma =1}^K \lambda_{\sigma }(1- p_{\sigma }^{-1}\tau_{\sigma }(\bW))}}$, we observe that for any $c \in \bbR$ and $\blambda \in \bbR^K$ it holds that $G(\blambda) = G(\blambda + c \bp)$, where $\bp = (p_1, \ldots, p_K)^\top \in \bbR^K$. Hence, since we are interested in any solution of the above optimization problem, we can define $(g^*, \blambda^*)$ as
\begin{align*}
     &g^*(\bw) = \ind{2\eta(\bw) - 1 \geq \sum_{\sigma =1}^K \lambda_{\sigma } p_{\sigma }^{-1}\tau_{\sigma }(\bw) }\enspace,\\
     &\blambda^* \in \argmin_{\blambda \in \bbR^K} \enscond{\Exp\parentsq{\abs{ 2\eta(\bW) - 1 - \sum_{\sigma=1}^K \lambda_{\sigma } p_{\sigma }^{-1}\tau_{\sigma }(\bW)}}}{\bar{\blambda} = 0}\enspace.\qedhere
\end{align*}

\end{proof}

\begin{lemma}
\label{lem:minimizer_exists}
Let Assumption~\ref{ass:continuity_unified} be satisfied, then
the mapping 
\begin{align}
\label{eq:unconstrained_min_dual}
\blambda \mapsto \Exp\parentsq{\parent{ 2\eta(\bW) - 1 + \sum_{\sigma =1}^K \lambda_{\sigma }(1- p_{\sigma }^{-1}\tau_{\sigma }(\bW))}_+}
\end{align}
attains its minimum.
\end{lemma}
\begin{proof}
In the end of the proof of Theorem~\ref{thm:optimal_DP_unified} we have show that minimization of \eqref{eq:unconstrained_min_dual} is equivalent to the minimization of
\begin{align*}
\blambda \mapsto \Exp\parentsq{\abs{ 2\eta(\bW) - 1 - \sum_{\sigma=1}^K \lambda_{\sigma } p_{\sigma }^{-1}\tau_{\sigma }(\bW)}}
\end{align*}
on the hyperplane $ \enscond{\blambda \in \bbR^K}{\bar{\blambda} = 0}$.
Thus, it is sufficient to show that 
\begin{align*}
    \min_{\blambda \in \bbR^K} \enscond{\Exp\parentsq{\abs{ 2\eta(\bW) - 1 - \sum_{\sigma=1}^K \lambda_{\sigma } p_{\sigma }^{-1}\tau_{\sigma }(\bW)}}}{\bar{\blambda} = 0}
\end{align*}
is attained.

It is clear that the mapping in question is convex on $\bbR^K$. Hence, it is sufficient to show that it is coercive~\cite[see e.g.][Proposition 11.15]{bauschke2017convex}.
It holds that
\begin{align}
    \label{eq:coercivity1}
    \Exp{\abs{ 2\eta(\bW) - 1 - \sum_{\sigma=1}^K \lambda_{\sigma } p_{\sigma }^{-1}\tau_{\sigma }(\bW)}} = \Exp\abs{\scalar{(\blambda / \bp, 1)}{(\bV, H)}} \enspace,
\end{align}
where we introduced the vector $\bV \triangleq (\tau_1(\bW), \ldots, \tau_K(\bW))$, $H \triangleq 1- 2\eta(\bW)$, and $(\blambda / \bp, 1) \triangleq (\lambda_1 / p_1, \ldots, \lambda_K / p_K, 1) \in \mathbb{R}^{K+1}$. Thus, in view of \eqref{eq:coercivity1}, by Markov's inequality, for any $\kappa > 0$ it holds that
\begin{align}
    \label{eq:existence_min_00}
    \Exp{\abs{ 2\eta(\bW) {-} 1 - \sum_{\sigma=1}^K \frac{\lambda_{\sigma }}{ p_{\sigma }}\tau_{\sigma }(\bW)}} \geq \kappa \|(\blambda / \bp, 1)\|\Prob(\abs{\scalar{(\blambda / \bp, 1)}{(\bV, H)}} > \kappa \|(\blambda / \bp, 1)\|)\enspace,
\end{align}
where $\|\cdot\|$ denotes the Euclidean norm.
Note that if we are able to show that for some $\kappa_0 > 0$, the right hand side of the above inequality is bounded away from zero, the proof of coercivity is concluded since $\|(\blambda / \bp, 1)\| \geq \min_{s \in [K]}\{ p_s^{-1}\}\|\blambda\|$.
To this end, let us introduce
\begin{align*}
    F(\bu, t) = \Prob(\abs{\scalar{\bu}{(\bV, H)}} \leq t)\enspace,
\end{align*}
for all $t \geq 0$ and $\bu \in \class{H}_0$ being defined as
\begin{align*}
    \class{H}_0 = \enscond{\bu \in \bbR^{K + 1}}{\|\bu\| = 1, \,\, \bu = (\lambda_1/p_1, \ldots, \lambda_K / p_k, 1) \text{ for some } \lambda_1 + \ldots + \lambda_K = 0}\enspace.
\end{align*}
By Assumption~\ref{ass:continuity_unified}, for any $\bu \in \class{H}_0$, the mapping $t \mapsto F(\bu, t)$ is continuous on $(0, +\infty)$ with $F(\bu, 0) = 0$ and $F(\bu, +\infty) = 1$. 
Furthermore, for any $\bu \in \class{H}_0, \bh \in \bbR^{K+1}$ such that $\bu + \bh \in \class{H}_0$ and for any $\delta > 0, t > 0$, we have thanks to triangle's inequality and monotonicity of $F(\bu, \cdot)$ 
\[
    F(\bu + \bh, t + \delta) \in \bigg[F(\bu, t + \delta - 2\|\bh\|),\, F(\bu, t + \delta + 2\|\bh\|)\bigg] \xrightarrow{\stackrel{\delta \longrightarrow 0}{\|\bh\| \longrightarrow 0}} F(\bu, t)\enspace,
\]
where the convergence follows from the assumed continuity of $F(\bu, \cdot)$. Thus, $(\bu, t) \mapsto F(\bu, t)$ is continuous. Since $\class{H}_0$ is compact, we have that
\begin{align*}
    G(t) \triangleq \sup_{\bu \in \class{H}_0} F(\bu, t)\enspace,
\end{align*}
is continuous on $[0, +\infty)$. Hence, the intermediate value theorem guarantees that there exists $\kappa_0 > 0$ such that
\begin{align*}
    G(\kappa_0) = 1 - \inf_{\lambda_1 + \ldots + \lambda_K = 0}\Prob(\abs{\scalar{(\blambda / \bp, 1)}{(\bV, H)}} > \kappa_0 \|(\blambda / \bp, 1)\|)  = \frac{1}{2}\enspace.
\end{align*}
In view of Eq.~\eqref{eq:existence_min_00}, we conclude.
\end{proof}

% \section{Omitted proofs results}

\section{Auxiliary results}

The first lemma ensures that under certain conditions, the denominator of the linear fractional performance measure is always positive.
\begin{lemma}
\label{lem:denominator_positive}
Assume that $\sfd_0 + \min \big\{\min\{\sfd_1,\, 0\} + \sfd_2,\, 0\big\} \geq 0$, then for any classifier $g : \class{X} \times [K] \to \{0, 1\}$
\begin{align*}
    &\sfd_0 + \sfd_1\Prob(Y = 1,\, g(\bX, S) = 1) + \sfd_2 \Prob(g(\bX, S) = 1) \geq 0\enspace.
\end{align*}
Furthermore, if $\sfd_0 + \min \big\{\min\{\sfd_1,\, 0\} + \sfd_2,\, 0\big\} > 0$, then the above inequality is strict.
\end{lemma}
\begin{proof}
Observe that
\begin{align*}
    \sfd_0 + \sfd_1\Prob(Y = 1,\, g(\bX, S) = 1) + \sfd_2 \Prob(g(\bX, S) = 1)
    &=
    \sfd_0 + \Exp[(\sfd_1 Y + \sfd_2)g(\bX, S)]\\
    &\geq
    \sfd_0 + \Exp[(\min\{\sfd_1,\, 0\} + \sfd_2)g(\bX, S)]\\
    &\geq
    \sfd_0 + \min \big\{\min\{\sfd_1,\, 0\} + \sfd_2,\, 0\big\}\\
    &\geq 0\enspace.
\end{align*}
The second claim follows the same lines.
\end{proof}

The second result gives a sufficient condition for positivity of the leading coefficient in Remark~\ref{rem:positive_excess}.
\begin{lemma}
\label{lem:sign_of_a_bad_boy}
Assume that $\sfd_0 + \min \big\{\min\{\sfd_1,\, 0\} + \sfd_2,\, 0\big\} \geq 0$ and either Eq.~\eqref{eq:condition1} or Eq.~\eqref{eq:condition2} is satisfied, then for any classifier $g \in \dom({\rm U}_{(\sfn, \sfd)})$
\begin{align*}
    \sfn_1 - \sfd_1{\rm U}_{(\sfn, \sfd)}(g) \geq 0\enspace.
\end{align*}
\end{lemma}
\begin{proof}
Observe that in both cases, by Lemma~\ref{lem:denominator_positive}, we have
\begin{align}
    \label{eq:sign_of_a_bad_boy}
    \sign(\sfn_1 - \sfd_1{\rm U}_{(\sfn, \sfd)}(g)) = \sign\bigg((\sfn_1\sfd_0 - \sfd_1\sfn_0) + (\sfn_1\sfd_2 - \sfd_1\sfn_2)\Exp[g(\bX, S)]\bigg)\enspace.
\end{align}
{\bf Case 1: } $\sfn_2\sfd_1 \neq \sfd_2\sfn_1$. In that case condition~\eqref{eq:condition1} implies that $\sfn_1\sfd_2 - \sfd_1\sfn_2 > 0$ and $\tfrac{\sfn_1\sfd_0 - \sfd_1\sfn_0}{\sfn_1\sfd_2 - \sfd_1\sfn_2} \geq 0$. In view of~\eqref{eq:sign_of_a_bad_boy} we conclude.\\
{\bf Case 2: } $\sfn_2\sfd_1 = \sfd_2\sfn_1$. The proof is immediate from~\eqref{eq:sign_of_a_bad_boy} and the first part of condition~\eqref{eq:condition2}.  
\end{proof}

The next lemma establishes an extended average stability property from~\citep{chzhen2020minimax}.
\begin{lemma}
\label{lem:f_star_eta_replacing}
Let Assumption~\ref{ass:continuity0} be satisfied, then
\begin{align*}
    \Exp[(f^*(\bX, S) - \eta(\bX, S))\ind{f^*(\bX, S) \geq \theta}] = 0\enspace,
\end{align*}
for all $\theta \in [0, 1]$.
\end{lemma}
\begin{proof}
Fix some $\theta \in [0, 1]$.
Introducing $T^*(\cdot) \coloneqq \parent{\sum_{\sigma = 1}^K p_{\sigma}F^{-1}_{\mu_{\sigma}(\eta)} }(\cdot)$, we recall that
\begin{align*}
    %f^*(\bx, s) = \parent{\sum_{\sigma = 1}^K p_{\sigma}F^{-1}_{\mu_{\sigma}(\eta)} } \circ F_{\mu_s(\eta)}(\eta(\bx, s)) = T^* \circ F_{\mu_s(\eta)}(\eta(\bx, s))\enspace.
    f^*(\bx, s) = T^* \circ F_{\mu_s(\eta)}(\eta(\bx, s))\enspace.
\end{align*}
Furthermore, since both $F_{\mu_S(\eta)}(\eta(\bX, S))$ and $(F_{\mu_S(\eta)}(\eta(\bX, S)) \mid S = s)$ are distributed uniformly on $(0, 1)$ under Assumption~\ref{ass:continuity0}, we can write
\begin{align*}
    \Exp[(&f^*(\bX, S) - \eta(\bX, S))g^*(\bX, S)]\\
    &=
    \Exp[T^*(U)\ind{T^*(U) \geq \theta}] -
    \sum_{s = 1}^K p_s\Exp[F_{\mu_s(\eta)}^{-1}(U)\ind{T^*(U) \geq \theta} \mid S = s]
    = 0\enspace.
    \qedhere
\end{align*}
\end{proof}

Finally, the last result relates the excess risk obtained in Lemma~\ref{lem:excess_score} with the expression presented in Remark~\ref{rem:positive_excess}.
\begin{lemma}
\label{lem:relating_bad_boy_to_bad_number}
Under the conditions of Lemma~\ref{lem:fixed_at_optim_LF}, we have
\begin{align*}
\sfn_1 - \sfd_1{\rm U}(g^*) = 
    \begin{dcases}
    \frac{\sfd_2\sfn_1 - \sfn_2\sfd_1}{\sfd_2 + \theta^*_{(\sfn, \sfd)}\sfd_1} &\text{if } \sfd_2\sfn_1 \neq \sfn_2\sfd_1\\
    \frac{\sfn_1\sfd_0 - \sfd_1\sfn_0}{\sfd_0 + \sfd_1\Exp(f^*(\bX, S) - \theta^*_{(\sfn, \sfd)})_+} &\text{if } \sfd_2\sfn_1 = \sfn_2\sfd_1
    \end{dcases}\enspace.
\end{align*}
\end{lemma}
\begin{proof}
We drop the subscript $(\sfn, \sfd)$ for compactness.\\
{\bf Case 1:} $\sfd_2\sfn_1 \neq \sfn_2\sfd_1$. Using the corresponding case of Lemma~\ref{lem:fixed_at_optim_LF} and solving it for $\theta^*$, we deduce that
\begin{align*}
    \theta^* = \frac{\sfn_2 - \sfd_2{\rm U}(g^*)}{\sfd_1{\rm U}(g^*) - \sfn_1}\enspace.
\end{align*}
Hence, from the above we deduce that
\begin{align*}
    \sfd_2 + \theta^*\sfd_1 = \frac{\sfd_1\sfn_2 - \sfd_2\sfn_1}{\sfd_1{\rm U}(g^*) - \sfn_1}\qquad \implies \qquad \frac{\sfd_2\sfn_1 - \sfn_2\sfd_1}{\sfd_2 + \theta^*\sfd_1} = \sfn_1 - \sfd_1{\rm U}(g^*)\enspace.
\end{align*}
{\bf Case 1:} $\sfd_2\sfn_1 = \sfn_2\sfd_1$. Again using the corresponding case of Lemma~\ref{lem:fixed_at_optim_LF} and solving it for $\Exp(f^*(\bX, S) - \theta^*)_+$, we deduce that
\begin{align*}
    \Exp(f^*(\bX, S) - \theta^*_{(\sfn, \sfd)})_+ = \frac{\sfd_0 {\rm U}(g^*) - \sfn_0}{\sfn_1 - \sfd_1{\rm U}(g^*)}\enspace.
\end{align*}
Hence, from the above we deduce that
\begin{align*}
    \sfd_0 + \sfd_1\Exp(f^*(\bX, S) - \theta^*)_+ = \frac{\sfd_0\sfn_1 - \sfd_1\sfn_0}{\sfn_1 - \sfd_1{\rm U}(g^*)} \implies\frac{\sfn_1\sfd_0 - \sfd_1\sfn_0}{\sfd_0 + \sfd_1\Exp(f^*(\bX, S) - \theta^*)_+} = \sfn_1 - \sfd_1{\rm U}(g^*)\enspace.
\end{align*}
The proof is concluded.
\end{proof}

%%%%%%%%%%%%%%%%%%%%%%%%%%%%%%%%%%%%%%%%%%%%%%%%%%%%%%%%%%%%%%%%%%%%%%%
%%%%%%%%%%%%%%%%%%%%%%%%%%%%%%%%%%%%%%%%%%%%%%%%%%%%%%%%%%%%%%%%%%%%%%%
\bibliography{biblio}
\bibliographystyle{apalike}
%%%%%%%%%%%%%%%%%%%%%%%%%%%%%%%%%%%%%%%%%%%%%%%%%%%%%%%%%%%%%%%%%%%%%%%
%%%%%%%%%%%%%%%%%%%%%%%%%%%%%%%%%%%%%%%%%%%%%%%%%%%%%%%%%%%%%%%%%%%%%%%

\end{document}